\documentclass{sinapisai}
\usepackage[numbers,square,sort&compress]{natbib}

\usepackage{tikz}
\usepackage{pgfplots}
\pgfplotsset{compat=1.18}
\usepackage{longtable}
\usepackage{multirow,colortbl} \usepackage{fontawesome5} \usetikzlibrary{positioning,arrows.meta,backgrounds,fit}
\usepackage[nameinlink,capitalize,noabbrev]{cleveref}

\title{\texorpdfstring{Toward Collective Model Intelligence:\\
  Characterizing and Coordinating Models}{Toward Collective Model Intelligence: Characterizing and Coordinating Models}}
\reporttitlesize{22}{26}
\shorttitle{Toward Collective Model Intelligence: Characterizing and Coordinating Models}
\reportteam{SinapisAI Team}
\affiliation{}
\date{2026-10}
\reportnumber{}
\reportversion{}
\authornote{}
\reportcontact{}
\reportnotice{}
\reportlinks{\rmfamily\fontsize{9}{12}\selectfont
  \hypersetup{urlcolor=ReportInk}\makebox[0pt][r]{\reportlink{{\color{black}\faGithub}\hspace{0.4em}SinapisAI}{https://github.com/SinapisAI}\hspace{2em}}\mbox{\reportlink{\raisebox{-0.17em}{\includegraphics[height=1.1em,
      trim=10bp 8bp 1499bp 5bp,clip]{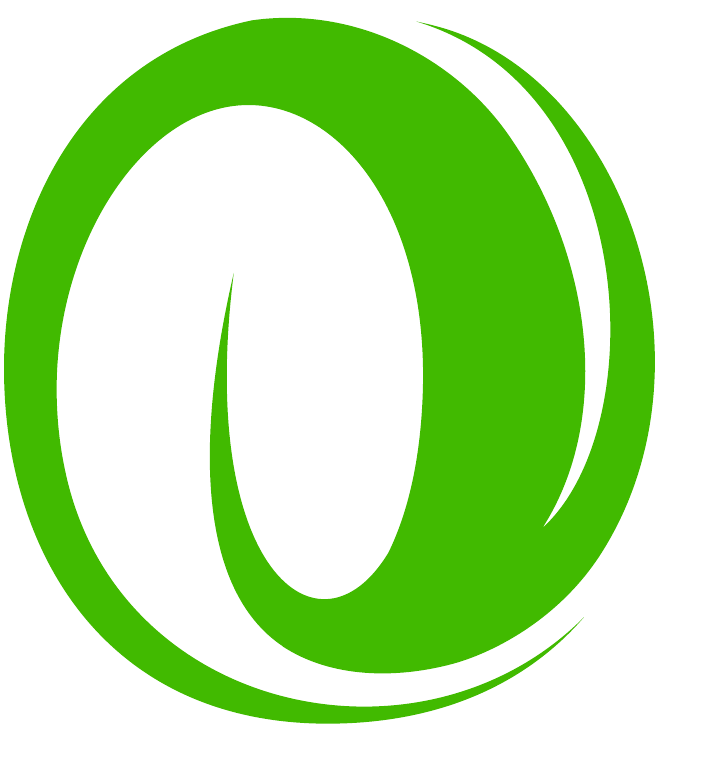}}\hspace{0.4em}https://sinapis.ai}{https://sinapis.ai}}\makebox[0pt][l]{\hspace{2em}\reportlink{\raisebox{-0.28em}{\includegraphics[height=1.4em]{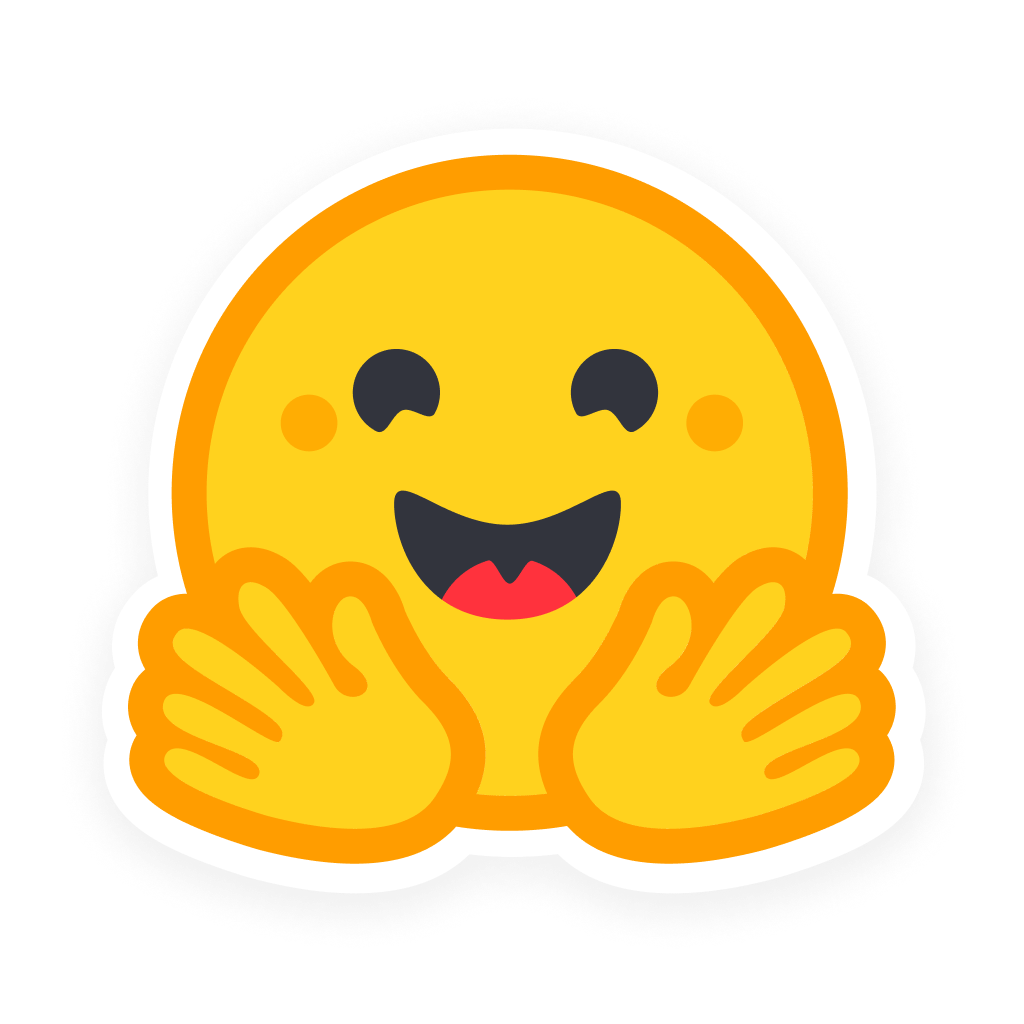}}\hspace{0.35em}SinapisAI}{https://huggingface.co/SinapisAI}}}

\begin{document}
\maketitle
\begin{abstract}
Independently developed AI models accumulate diverse and complementary
capabilities, but a growing model collection does not automatically realize
\emph{Collective AI}. Addressing this gap requires
\emph{understanding model capabilities with limited evaluation,
organizing complementary models to collaborate on diverse tasks,
and executing model calls efficiently under resource constraints}.
We present a systematic approach to pursuing
\emph{Collective Model Intelligence}
through three connected components.
First, Model Eval constructs evidence-based capability profiles by combining query-level performance estimation with cross-domain prediction, reducing the need for exhaustive evaluation. Second, Mio (Models in One) organizes complementary model candidates and supports task-conditioned, cost-aware selection and collaboration as task requirements evolve. Third, Mos (Model Serving OS) separates model selection from execution and provides endpoint scheduling and shared model deployment and distribution.
Experiments show lower capability-estimation errors under limited evaluation. Across nine public benchmarks, Mio 1.0 Lite achieves the highest average score and lowest average cost among compared methods, with per-request costs 73.7--97.5\% lower than those of the best-scoring alternatives. A prewarmed shared cache accelerates model-file retrieval by up to \(23.3\times\).
This work lays a foundation for Collective Model Intelligence by characterizing and coordinating the diverse capabilities of an expanding model universe.
\end{abstract}

\begingroup
\definecolor{EvidenceMio}{HTML}{D44773}
\definecolor{EvidenceTagline}{HTML}{7A8278}
\newcommand{\evidencetitle}[2]{\noindent{\fontsize{12}{14}\selectfont\bfseries\color{BrandInk}#1}\hfill
  {\fontsize{8}{10}\selectfont\color{EvidenceTagline}#2}\par}
\newcommand{\evidenceauxcaption}[1]{\vspace{1pt}\raggedright
  {\fontsize{8.5}{11}\selectfont\bfseries\color{BrandInk}\strut#1\par}}
\begingroup
\setlength{\parskip}{0pt}
\vspace{11pt}
\vfill
\noindent\begin{minipage}[t]{\linewidth}
\hypertarget{mio-frontpage-results}{}
\evidencetitle{Mio 1.0 Lite}{Collaboration among cost-efficient models from China.}
\vspace{1pt}
\noindent\includegraphics[width=\linewidth]{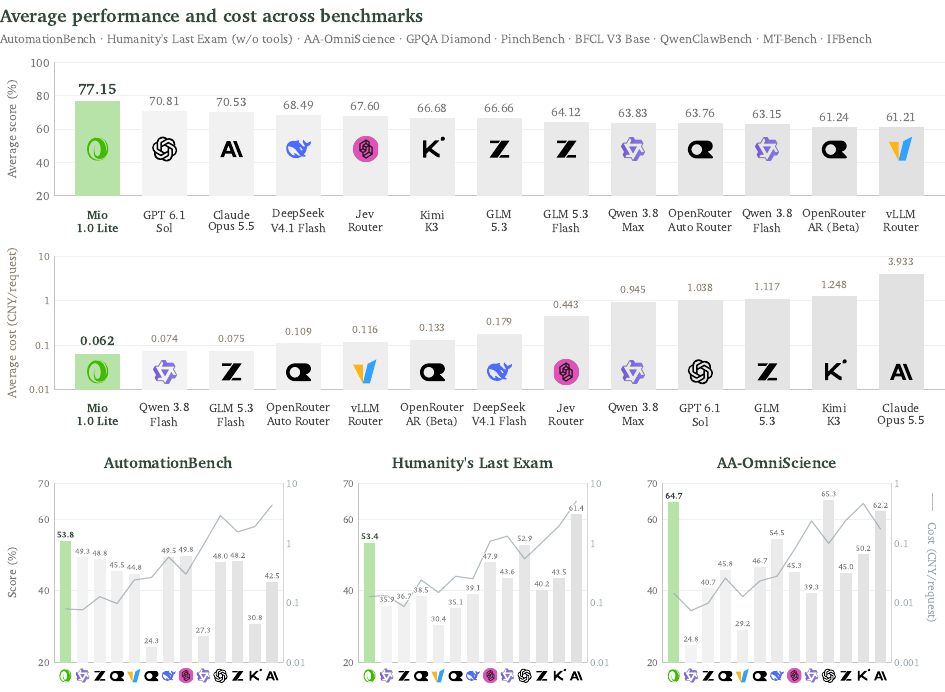}
\par\vspace{2pt}
\evidenceauxcaption
  {Mio 1.0 Lite achieves the highest average score and lowest average cost, with per-request costs 73.7--97.5\% lower than those of the best-scoring alternatives across benchmarks
   {\normalfont\color{ReportMuted}(full results across all benchmarks: see
    \ifcsdef{r@sec:model-selection-collaboration-performance}{\hyperref[sec:model-selection-collaboration-performance]{\underline{Section~\ref*{sec:model-selection-collaboration-performance}}}}{Section~6.1})}.}
\end{minipage}
\par\vfill
\endgroup
\endgroup

\clearpage
\begingroup
\section{From Individual Models to Collective AI}
\label{sec:introduction}
\subsection{The Expanding Model Universe}
\label{sec:expanding-model-universe}
AI resources are distributed across organizations and environments.
Datasets, trained models, domain expertise, and computation cannot always
be brought together because of privacy, ownership, and communication
constraints. Independent training adds new models, while continued
adaptation changes the capabilities of existing ones. Together, these
processes sustain an expanding and evolving \emph{model universe}: a
population of models developed or adapted across organizations, datasets,
and task settings. Models differ in architecture,
modality, and specialization; even those derived from the same foundation
model can develop distinct strengths through adaptation.

These models embody knowledge and skills accumulated across developers,
datasets, and tasks. Their diversity creates opportunities to reuse
complementary capabilities
\citep{zhou2026modelmanagement}.
A system can realize additional value by understanding these capabilities,
organizing complementary models, and coordinating their contributions to tasks.

We pursue this opportunity through \emph{Collective AI}: coordinating
distributed data, computational resources, and intelligent capabilities to
learn, reason, and act at the system level
\citep{jordan2026collectivist,genewein2026agitoasi}.
We consider three complementary levels of coordination. At the \emph{data}
level, Collective Data Intelligence develops model capabilities by learning
from diverse data sources, accumulated experience, and feedback.
Many familiar learning practices can be viewed through this lens,
including large-scale language and multimodal pretraining
\citep{brown2020gpt3,radford2021clip}, learning from feedback through
reinforcement learning \citep{guo2025deepseekr1}, and federated learning
\citep{mcmahan2017federated}. At the
\emph{computation} level, distributed training and inference coordinate work
across computing resources. At the \emph{capability} level, coordination
combines complementary knowledge, skills, and functions supplied by models,
agents, and tools to accomplish tasks.

Realizing these benefits requires effective coordination.
At the capability level, for example, a model with a high average score may
largely duplicate strengths already present in the collection, while a
specialist may contribute a missing capability. A useful collective must
identify and use these differences to improve outcomes or resource
efficiency, accounting for coordination costs.

Trained models offer a practical starting point for capability-level
coordination because they embody accumulated learning. We envision a
collective that can grow to incorporate diverse models, including
specialists for different domains and tasks. In this vision, the collective
also uses experience and its own capabilities to improve how it is
organized and coordinated.
Work on continual learning under resource constraints
\citep{javed2024bigworld} and learning from interaction
\citep{silver2025experience} motivates the emphasis on adaptation and
experience.

\subsection{Collective Model Intelligence for Collective AI}
\label{sec:model-intelligence-challenge}
At SinapisAI, we focus on \emph{Collective Model Intelligence}:
understanding, reusing, and coordinating capabilities across an evolving
population of independently developed models to improve task outcomes and
resource efficiency. This complements Collective Data Intelligence by reusing
capabilities encoded in trained models, with data, evaluation, and
feedback guiding coordination.

Reusing accumulated capabilities requires information about where a model
is applicable and what it can contribute. The learnware paradigm addresses
this need by pairing trained models with specifications that help users
identify and reuse suitable capabilities
\citep{zhou2016learnware,zhou2024smallmodels,zhou2026modelmanagement}.

For example, a task involving visual interpretation, reasoning, and tool
use may benefit from different models at different stages. Other tasks may
call for selecting among models with different strengths and resource
requirements. These possibilities depend on identifying and using model
complementarity in the context of the task.

The central challenge is to infer useful model complementarity from limited
evidence and determine when coordinating these capabilities yields a net
gain under task and resource constraints. As models, tasks, and service
conditions change, these decisions must adapt. As an initial step toward
this broader vision, this report addresses three foundational questions:
\begin{enumerate}
\item \textbf{How can capabilities be characterized with limited evaluation?}
Capability characterization must estimate model performance across domains
and queries from sparse observations and identify complementary model strengths.
\item \textbf{How can models be organized for selection and collaboration?}
Organization must preserve useful capability diversity, while task-dependent
selection and collaboration must adapt model use to evolving needs and remaining
cost and latency budgets.
\item \textbf{How can inference requests be executed efficiently under service constraints?}
Scheduling must account for changing load and availability across
heterogeneous endpoints, while efficient model distribution supports
self-managed runtimes.
\end{enumerate}

Capability evidence informs which models are retained and used, while
execution conditions determine whether a choice can be carried out within
the task's constraints. Evaluation and task feedback can revise this
evidence and guide subsequent decisions. These connections motivate a
system that links capability characterization, model organization and use,
and efficient execution.

\subsection{A System for Collective Model Intelligence}
\label{sec:sinapisai-system}
We address these three questions through a system that connects
Model Eval for capability characterization, Mio for model organization,
selection and collaboration, and Mos for shared execution. Model Space
provides the shared inventory linking models, capability evidence, and
execution resources (\cref{sec:system-framework}). These components form an
initial foundation within the broader vision illustrated in
\cref{fig:collective-vision}.

\paragraph{Characterizing capabilities.}
Model Eval uses historical evidence and limited additional evaluation to
construct reusable capability profiles. It combines score estimation from
limited query observations with prediction across domains, providing
capability evidence for model organization and use
(\cref{sec:capability-characterization}).

\paragraph{Organizing models for selection and collaboration.}
Mio (Models in One) is designed to organize a compact pool of complementary
models for task-dependent selection and collaboration. Capability evidence
guides pool construction, while task requirements and remaining cost and
latency budgets guide model use. Selection assigns a suitable model to the
current task, while collaboration lets complementary models contribute at
different stages. Mio uses earlier results and the remaining cost and
latency budgets to decide which model to call next
(\cref{sec:mio}).

\paragraph{Executing inference requests efficiently.}
Mos (Model Serving OS) provides shared execution for capability evaluation
and application calls. It schedules calls across compatible endpoints and
coordinates model-file distribution for self-managed deployment, preserving
the model choice made by Model Eval, Mio, or the caller
(\cref{sec:model-serving}).

\Cref{sec:empirical-evaluation} reports component-level evidence of improved
score estimation under limited evaluation. Across nine public benchmarks,
Mio 1.0 Lite achieves the highest average score and lowest average
per-request cost among the compared methods. A prewarmed shared cache also
accelerates model-file retrieval.

\begin{figure}[htbp]
\centering
\includegraphics[width=\linewidth]{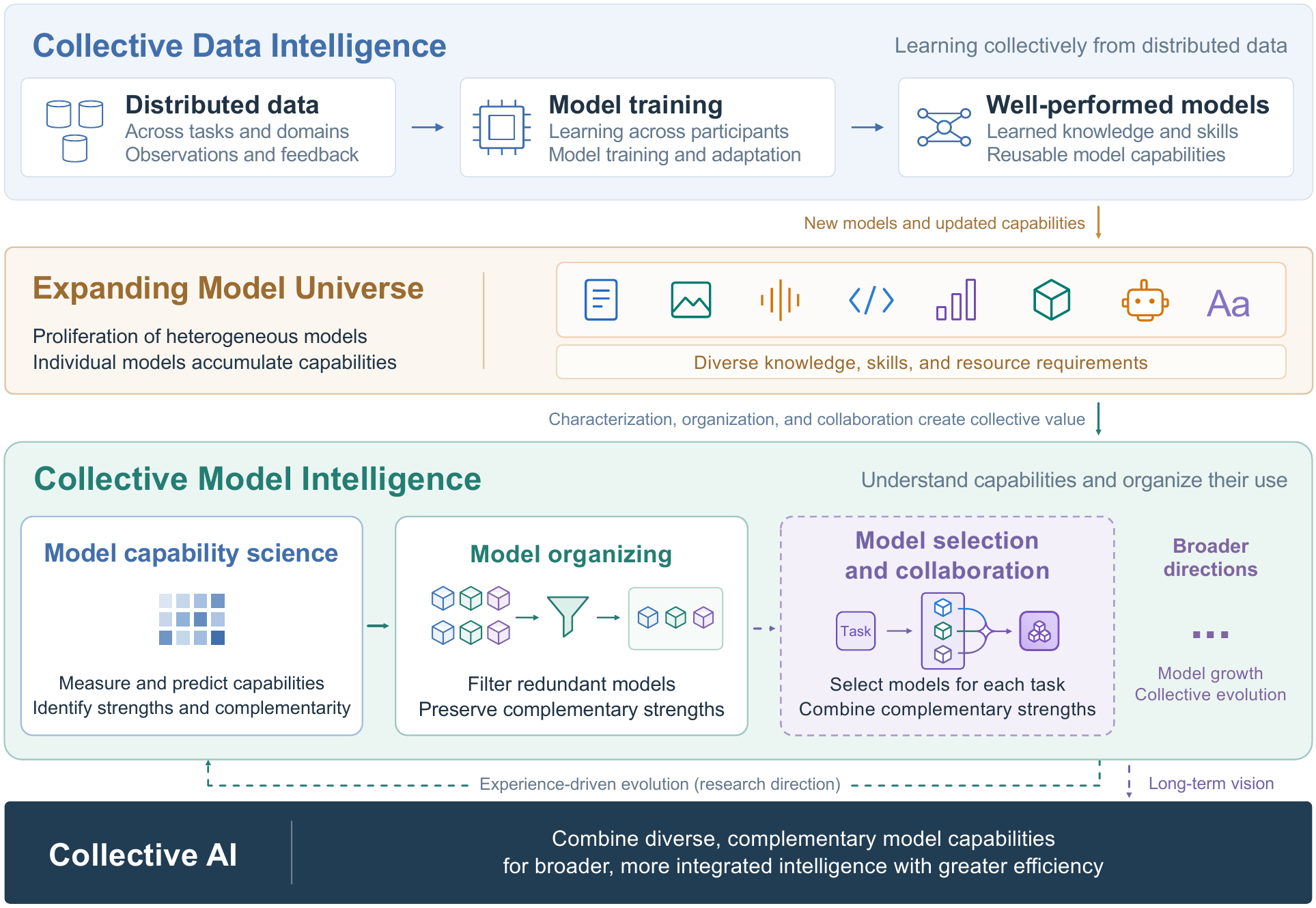}
\caption{Individual models accumulate knowledge and skills across
developers and domains. Collective Model Intelligence seeks additional value
by characterizing, organizing, and coordinating these capabilities. The
illustrated functions provide a foundation within a broader vision that
includes growth of the model population and experience-driven collective
evolution.}
\label{fig:collective-vision}
\end{figure}

\section{System Framework for Collective Model Intelligence}
\label{sec:system-framework}
\label{sec:infrastructure-design}
\label{sec:system-overview}
This section describes the architecture and operating contracts of our
Collective Model Intelligence Framework. We first introduce its components,
shared notation, and cross-component controls, then follow a model into
the candidate pool and a request through execution. Later sections detail
the component implementations and coordination mechanisms.

\subsection{Architecture Overview}
\label{sec:architecture-overview}

\begin{figure}[!htb]
\centering
\includegraphics[width=\linewidth]{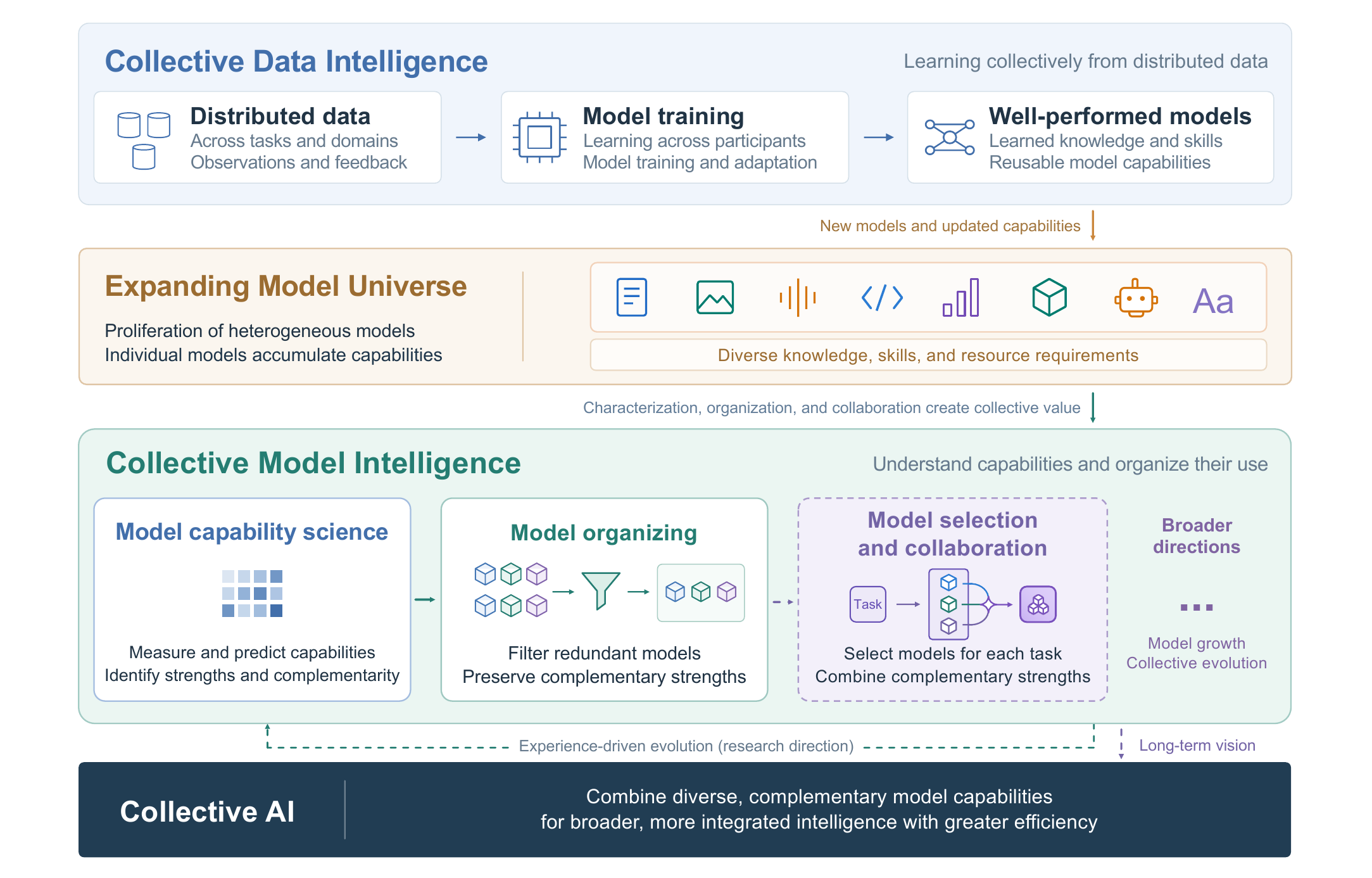}
\caption{Collective Model Intelligence Framework architecture.}
\label{fig:overview}
\end{figure}

\Cref{fig:overview} places the framework between AI Applications and the
underlying model services. Model Eval supplies query- and domain-level
capability profiles to Mio, which organizes candidates and coordinates their
use. Model Space provides shared model identities and resource references;
Mos uses these references to execute calls from Model Eval and Mio.
Capability evidence informs model choice, while current service observations
inform endpoint execution. Both kinds of records link to the same model
identities, allowing evaluation and application calls to share resources.

Below the framework, Mos invokes provider-managed External Model APIs or
manages the Self-Managed Model Runtime Hub. AI Cloud Infrastructure supplies
servers, GPU resources, storage, and networking in public or private clouds.
The Self-Managed Model Runtime Hub runs on these resources; external
endpoint deployment and operation remain the provider's responsibility.

We write a managed model as $(f_i,\Omega_i)$, following the
model--specification convention of \citet{zhou2026modelmanagement}.
Here $f_i$ identifies a model and $\Omega_i$ is the capability profile
published by Model Eval. This profile implements the model specification
through scoped scores, coverage, provenance, and uncertainty when available.
It is distinct from model weights and current endpoint state.
\Cref{sec:capability-characterization} explains how profiles are constructed.

The full inventory is $\mathcal F=\{f_1,\ldots,f_M\}$, and Mio publishes
a candidate pool $\mathcal C\subseteq\mathcal F$. Model $f_i$ can have several
endpoints, denoted by $\mathcal E(f_i)$; request-specific feasibility checks
restrict the pool before selection. A system snapshot records model,
profile, candidate-pool, and service versions. Formulas suppress version
indices, and comparisons use the declared snapshot; changes to the inventory
or scoring protocol require a new one. Live service observations complement
the snapshot during request execution. For example, two compatible endpoints
for the same model revision may share capability evidence but differ in price
or latency. Mos chooses between them using current service conditions.

Capability profiles organize evidence in a domain--query hierarchy shared
by Model Eval and Mio. A \emph{query} is a task instance evaluated under a
declared scoring protocol. An \emph{evaluation domain} groups queries with
a fixed population, score normalization, and aggregation rule. A benchmark
or a declared query cluster can define a domain. Profiles retain source and
query identifiers, along with the scoring protocol used.

Throughout the report, $d_j$ denotes an evaluation domain and
$\mathcal Q_j$ its declared query population. The score $s_{ij}$ is model
$f_i$'s normalized aggregate score on that domain, and $y_i(x)$ is its score
on query or request $x$. A hat marks an estimate or a completed score that
preserves observations. Monetary cost is $c$ and latency is $\ell$.
Evaluation budget $B$ and consumption $b$ use the study's declared unit;
request execution uses cost budget $C$ and latency allowance $D$.

\subsection{Core System Modules}
\label{sec:core-system-modules}
\paragraph{Model Eval.}
\label{sec:model-eval}
\label{sec:model-capability-evaluation}
Model Eval implements capability characterization using historical evidence
and budgeted evaluation through Mos. It publishes versioned profiles linked
to model and endpoint identities, the reference snapshot, and evaluation
configuration. Measured scores and predictions remain distinguishable, with
coverage and provenance supporting comparisons and uncertainty recorded when
available. Profiles retain query populations and scoring protocols so that
Mio can compare capability evidence under compatible evaluation conditions.
Characterization can be requested independently of candidate-pool
admission. New evidence or model revisions trigger profile refresh under
the lifecycle controls below. \Cref{sec:capability-characterization}
develops the evaluation framework and prediction methods.

\paragraph{Mio: Models in One.}
Mio contains Organization and Collaboration. Organization uses domain
coverage, query-level complementarity, and service compatibility to maintain
a compact candidate pool. Pool maintenance preserves complementary successes
and economical options;
removing a candidate requires a retained model that meets the declared
replacement-loss tolerances and service requirements.
Model selection and collaboration adapt to the bounded task state, current
service context, and remaining budgets as tasks progress. Mio passes the
selected identity to Mos for endpoint scheduling. \Cref{sec:mio} develops
model organization, selection and collaboration.

\paragraph{Model Space.}
\label{sec:modelspace}
\label{sec:model-management-runtime}
Model Space maintains the full inventory of API-accessible and downloadable
models, including resources outside Mio's active pool. Versioned identities
link sources, interfaces, artifacts or API access configuration, deployment
records, endpoints, and capability evidence. Family names group versions;
additional endpoints for a known model need not create new model identities.
Model Eval, Mio, and Mos link their evidence and decision records to these
identities. Discovery exposes available interfaces and compatible evidence,
while task suitability is assessed through characterization and selection.

\paragraph{Mos: Model Serving OS.}
Mos assigns evaluation, designated-model, and Mio-selected requests to
compatible inference endpoints. Its scheduling service uses current load,
price, and availability observations; its distribution and deployment
service prepares downloadable models for the runtime hub; and its
metrics/profiler records attempts, usage, and outcomes. External APIs and
self-managed runtimes share the invocation contract, with deployment and
model-file distribution applying to the self-managed path. Endpoint support
records specify modalities, streaming behavior, and tool interfaces.
\Cref{sec:slo-aware-scheduling} describes SLO-aware endpoint scheduling
and bounded recovery;
\cref{sec:efficient-distribution} develops model distribution for
self-managed model deployment.

\paragraph{Self-Managed Model Runtime Hub.}
The runtime hub provides the self-managed execution path below the
framework. Inference backends run model calls, instance lifecycle management
controls deployment readiness and availability, and a shared model-file
cache supports reuse. Mos prepares and invokes these deployments using
compute and storage from AI Cloud Infrastructure.

\paragraph{Cross-component controls.}
Lifecycle management, privacy, and security apply across the components
shown in \cref{fig:overview}.

\paragraph{Lifecycle updates.}
\label{sec:model-ecosystem-evolution}
\label{sec:runtime-lifecycle}
Lifecycle management connects resource, evidence, and candidate versions.
A new artifact revision requires readiness checks and characterization;
updated profiles can trigger candidate review under
\cref{sec:capability-guided-organizing}. Endpoint outages change request
eligibility, while price changes update cost context and can invalidate
pool replacements justified by a cost tier. These service events do not
themselves change measured capability. Earlier profiles and decision
snapshots remain available for interpreting past calls, and models removed
from Mio's pool retain their resources and history in Model Space.

\paragraph{Privacy and security.}
The design requires access checks before resource exposure or invocation,
controlled credential references, and declared storage and transmission
policies for inputs, outputs, artifacts, and logs. Credentials remain outside
capability profiles. Published evaluation summaries must preserve
private-task restrictions, and diagnostic traces
must limit sensitive content. Downloadable models require artifact
validation and runtime isolation; external services require explicit
provider data-handling policies. Deployment-specific security policies cover
authentication, isolation, and data retention.
\par

\subsection{Model Organizing Workflow}
\label{sec:model-onboarding}
\begin{figure}[!htb]
\centering
\includegraphics[width=\linewidth]{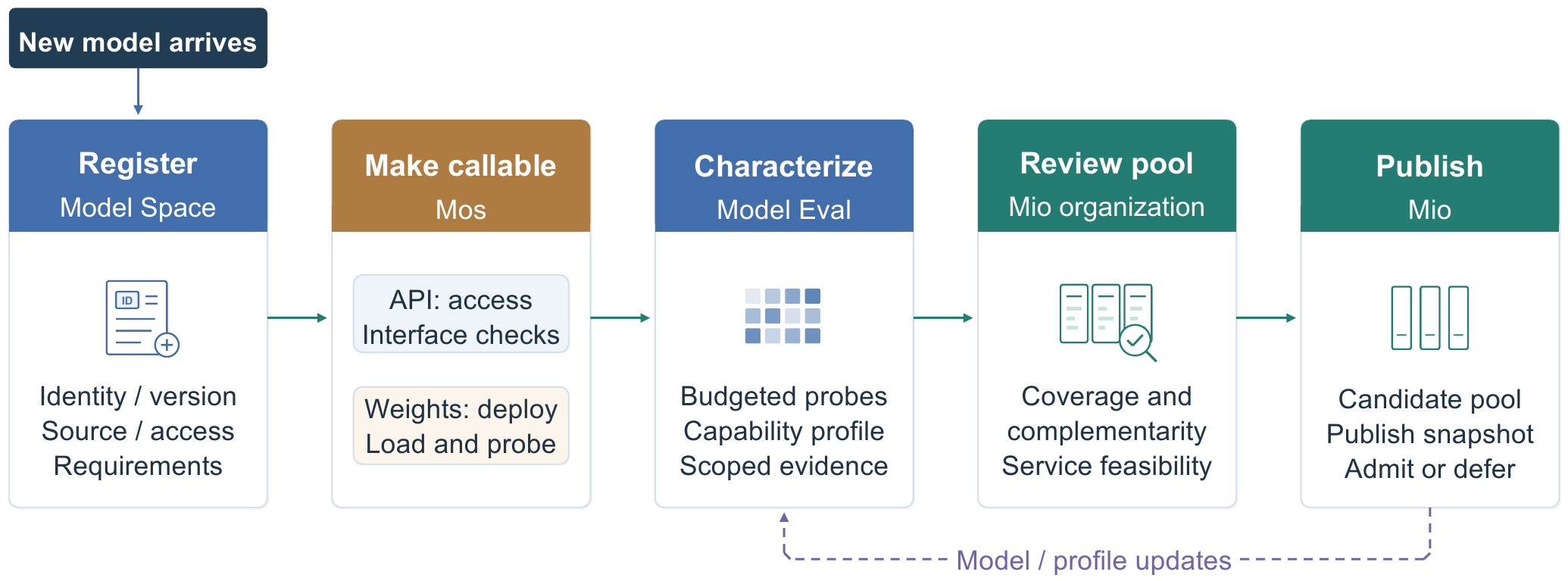}
\caption{Model organization workflow: register in Model Space, make callable
through Mos, characterize through Model Eval, and review and publish the
candidate pool through Mio.}
\label{fig:model-organizing-workflow}
\end{figure}
\paragraph{Register.} The path in \cref{fig:model-organizing-workflow} starts
with a resource declaration and resolved version in Model Space.
Downloadable models link source artifacts to runtime requirements; API
models link provider identifiers to access configuration and constraints.
Version resolution associates provider aliases with the model revision
described by existing capability evidence. Identity, artifact, and interface
validation determine readiness for the next stage.

\paragraph{Make callable.} Mos prepares the appropriate execution path.
Downloadable models undergo acquisition, deployment, loading, and readiness
probes in the runtime hub. API models undergo access and interface checks.
Access and readiness checks verify the declared interface and execution
configuration before exposing the resource for evaluation or user calls.
Task capability is assessed through Model Eval.

\paragraph{Characterize.} Model Eval executes budgeted probes through Mos and
scores their outcomes. It binds observations to model and execution versions
and publishes a profile with measured and inferred fields separated.

\paragraph{Review pool.} Mio checks the newcomer's coverage, complementarity,
and service feasibility alongside existing candidates. It may admit the
model, defer admission, or replace redundant candidates using the criteria
in \cref{sec:capability-guided-organizing}.

\paragraph{Publish.} Mio publishes the updated pool snapshot and maintains a
record of all admissions, deferrals, and replacements.

\begin{table}[htbp]
\centering
\caption{Model resource states and their supporting records at a recorded snapshot.}
\label{tab:resource-states}
\small
\begin{tabularx}{\linewidth}{@{}p{0.19\linewidth}>{\raggedright\arraybackslash}X>{\raggedright\arraybackslash}X@{}}
\toprule
State & Supporting record & Consequence \\
\midrule
Registered & Identity, source, version and access configuration & Discoverable in Model Space. \\
Callable & Compatible endpoint and readiness evidence & Available for execution under its access policy. \\
Characterized & Published, scoped capability evidence & Can support comparisons or capability-based decisions. \\
Mio candidate & Published pool snapshot and selection support & Included in the pool used for request-level selection. \\
\bottomrule
\end{tabularx}
\end{table}

The states in \cref{tab:resource-states} need not advance or regress
together: a characterized model can become temporarily unavailable, and a
callable model can remain uncharacterized. A new model revision requires
fresh checks and evidence; its predecessor's profile continues to describe
the earlier revision. Each request is checked against current endpoint
conditions and access constraints.

\subsection{Model Serving Workflow}
\label{sec:request-execution}
\begin{figure}[!htb]
\centering
\includegraphics[width=\linewidth]{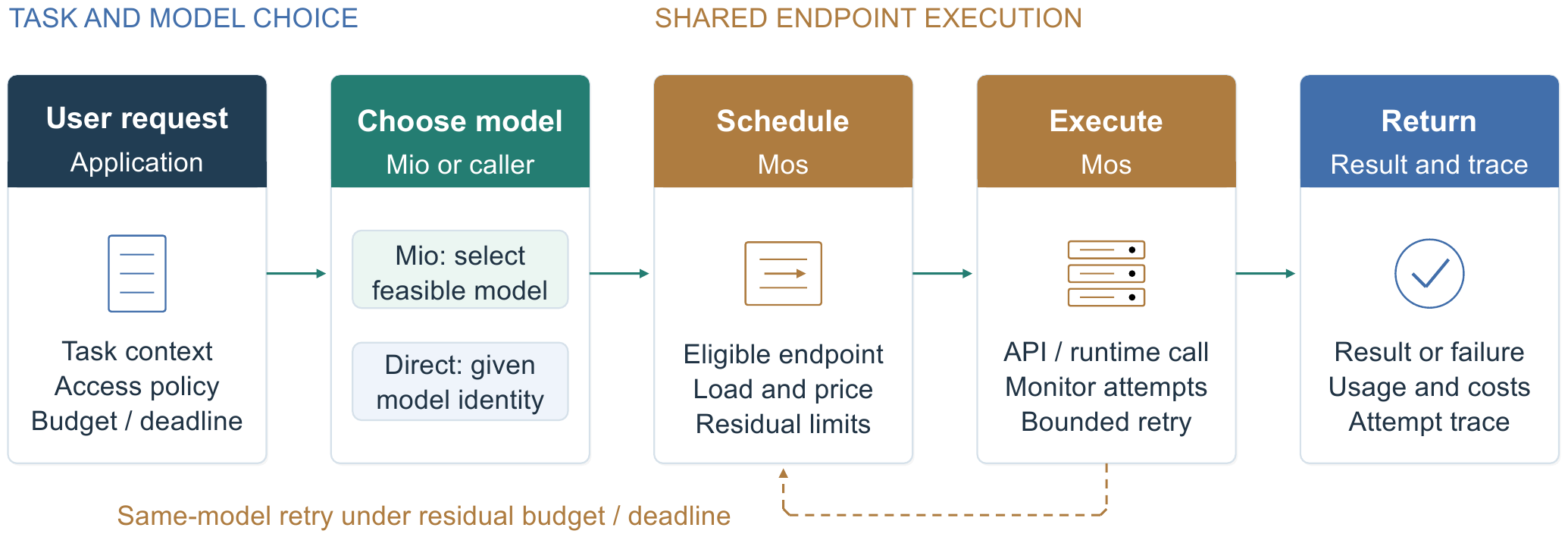}
\caption{Model serving workflow: select a model through Mio or the caller,
schedule and execute through Mos, and return results. The retry design preserves
model identity and residual constraints.}
\label{fig:model-serving-workflow}
\end{figure}
\paragraph{User request.} The path in \cref{fig:model-serving-workflow} begins
with task context, input requirements, a monetary budget, a completion
deadline, and an access policy. For a coding task, this context may include
file references, error messages, and earlier test results.

\paragraph{Choose model.} The caller supplies a designated model identity, or
Mio selects one from the task state and records the candidate, capability,
and policy versions consulted. Both paths use the same invocation contract;
endpoint selection subsequently resolves a service for that identity.

\paragraph{Schedule.} Mos validates required features and assigns an endpoint
using current load, price, and service observations. The SLO-aware scheduling
design checks feasibility against the remaining budget and deadline
(\cref{sec:slo-aware-scheduling}). The handoff carries residual allowances
after model selection and upstream work, preserving the request's spending
limit and clock origin.

\paragraph{Execute.} Mos invokes the selected API endpoint or runtime instance
and monitors the attempt. Under the recovery policy described here, a permitted
retry returns to scheduling for the same model version within the residual
constraints. It may use a compatible alternative endpoint after renewed
feasibility checks. Changing models returns control to Mio or the caller.
Every attempt retains its elapsed time and possible charges.

\paragraph{Return.} Mos returns the result or an explicit failure, usage,
available cost information, and a trace linking the chosen model and
endpoint to the attempt outcomes.

Operational feedback updates service context. Model Eval updates capability
evidence using scored outcomes, provenance checks, and evidence review.
In multi-step tasks, returned results and resource
usage update the task state and
remaining budgets before the next Mio decision
(\cref{sec:model-identification}). Traces preserve the snapshot used for
each choice.

\section{Model Capability Characterization}
\label{sec:capability-characterization}
\label{sec:capability-representation}
\label{sec:efficient-characterization}
\label{sec:efficient-capability-characterization}
\label{sec:efficient-evaluation-capability-representation}
\label{sec:evaluation-setting}
Model capability characterization aims to describe what a model can do,
where it tends to fail, and under which task conditions its capabilities
are useful. It produces a reusable representation for model comparison,
organization, selection and collaboration. In our framework, this process constructs and updates the
\emph{capability profile} $\Omega_i$ introduced in
\cref{sec:architecture-overview}; Model Eval is the system module that
collects evidence, applies the characterization methods, and publishes
the profile. Evaluation collects and scores model responses;
characterization combines this evidence with performance predictions
into profiles for model organization, selection and collaboration. The profile serves
the role of a model specification in the learnware paradigm.

Such a specification can take several forms. Semantic descriptions can
record intended tasks, modalities, and application contexts. Statistical
specifications can summarize properties of the data associated with a
model, as in reduced kernel mean embedding specifications
\citep{zhou2024smallmodels,tan2024beimingwu}. Evaluation-based
representations record model behavior on declared tasks. Together, these
forms describe task identity, distributional
compatibility, and demonstrated performance.

This section focuses on large language models (LLMs) and their multimodal
extensions, whose capabilities
span diverse tasks and can vary substantially across queries within the
same task. For such models, a broad semantic label provides limited
resolution for comparing strengths and failures, while hosted access may
provide neither training data nor internal parameters. Evaluation through
a model's callable interface supplies behavioral evidence under specified
task and scoring conditions \citep{liu2026constructive}.
We therefore construct evaluation-based
capability profiles using the domain--query hierarchy defined in
\cref{sec:architecture-overview}. Each profile combines domain scores with
query-level outcomes, uncertainty, and coverage. Domain scores describe
broad performance;
query-level evidence reveals complementary successes hidden by aggregate rankings.

For example, two models may have similar coding scores yet excel on
different queries, such as code generation versus debugging. Query-level
prediction estimates unqueried outcomes from a few scored queries in selected
\emph{anchor domains}, yielding aggregate \emph{anchor scores}. Cross-domain
prediction uses these scores and historical model-domain relationships to
estimate performance in untested domains, such as data analysis.

Constructing these profiles creates an evaluation challenge. Testing every
model on every query requires substantial inference and scoring resources
\citep{polo2024tinybenchmarks}, and historical evaluations cover models
and tasks unevenly. The objective is to characterize model $f_*$ across
evaluation domains using historical evidence and limited additional query
evaluations. Our approach combines two prediction
stages: query-level predictions estimate scores on selected anchor domains,
and relationships across models and domains reconstruct the remaining
scores. A multi-round procedure allocates further evaluation between
broadening domain coverage and refining anchor-score estimates, building the
profile $\Omega_*$ used by Mio. Model Eval publishes the profile with
observations, predictions, uncertainty, and provenance distinguished;
renewed evaluation updates this evidence as models and tasks change.
Characterization covers the declared model inventory and can be used
independently of Mio's candidate-pool admission and request-selection decisions.

The domain--query hierarchy and score notation follow the shared
evaluation organization in \cref{sec:architecture-overview}.
Let $\mathcal D=\{d_1,\ldots,d_J\}$ be the fixed inventory of $J$ domains.
The domain and query scores are organized into the partially observed
model-domain matrix
$S=(s_{ij})\in[0,1]^{M\times J}$ and the per-domain model--query matrices
$Y_j=(y_i(x))_{i,\,x\in\mathcal Q_j}$.
Historical observations are indexed by $\mathcal O_d$ for model-domain
pairs $(i,j)$ and $\mathcal O_q$ for model--query pairs $(i,x)$ used to fit
the query-response model. Different models may answer different query
subsets. Observation masks distinguish measured scores from unmeasured
entries and record each domain's query coverage.

The model-domain matrix $S$ is highly sparse because each historical model
has measured scores on only a small subset of domains. Among models with
historical evaluations in domain $d_j$, the corresponding rows of $Y_j$
are comparatively less sparse: evaluations record outcomes for many
queries in $\mathcal Q_j$. Query coverage can still be incomplete and
uneven across models. This difference motivates prediction within domains
followed by reconstruction across domains.

Within an evaluation budget, the objective is to estimate
$\mathbf s_*=(s_{*1},\ldots,s_{*J})$ together with uncertainty and coverage.
The model and domain inventories, query populations, scoring rules, and
historical fitting evidence remain fixed during a run; only the available
evaluation evidence for $f_*$ grows. Scores from incompatible populations
or protocols measure different quantities.

\subsection{Multi-Round Hierarchical Evaluation Framework}
\label{sec:hierarchical-anchor-estimation}
The framework in \cref{fig:model-evaluation} alternates between collecting
query evidence and updating $\Omega_*$. Each round selects queries
from the domain-query hierarchy, executes and scores them,
estimates anchor scores, and predicts the full domain-score vector.
The remaining budget can broaden domain coverage or evaluate more queries
in an existing anchor domain to improve precision. Historical
components of both estimators are fitted before profiling $f_*$.

\begin{figure}[htbp]
\centering
\includegraphics[width=\linewidth]{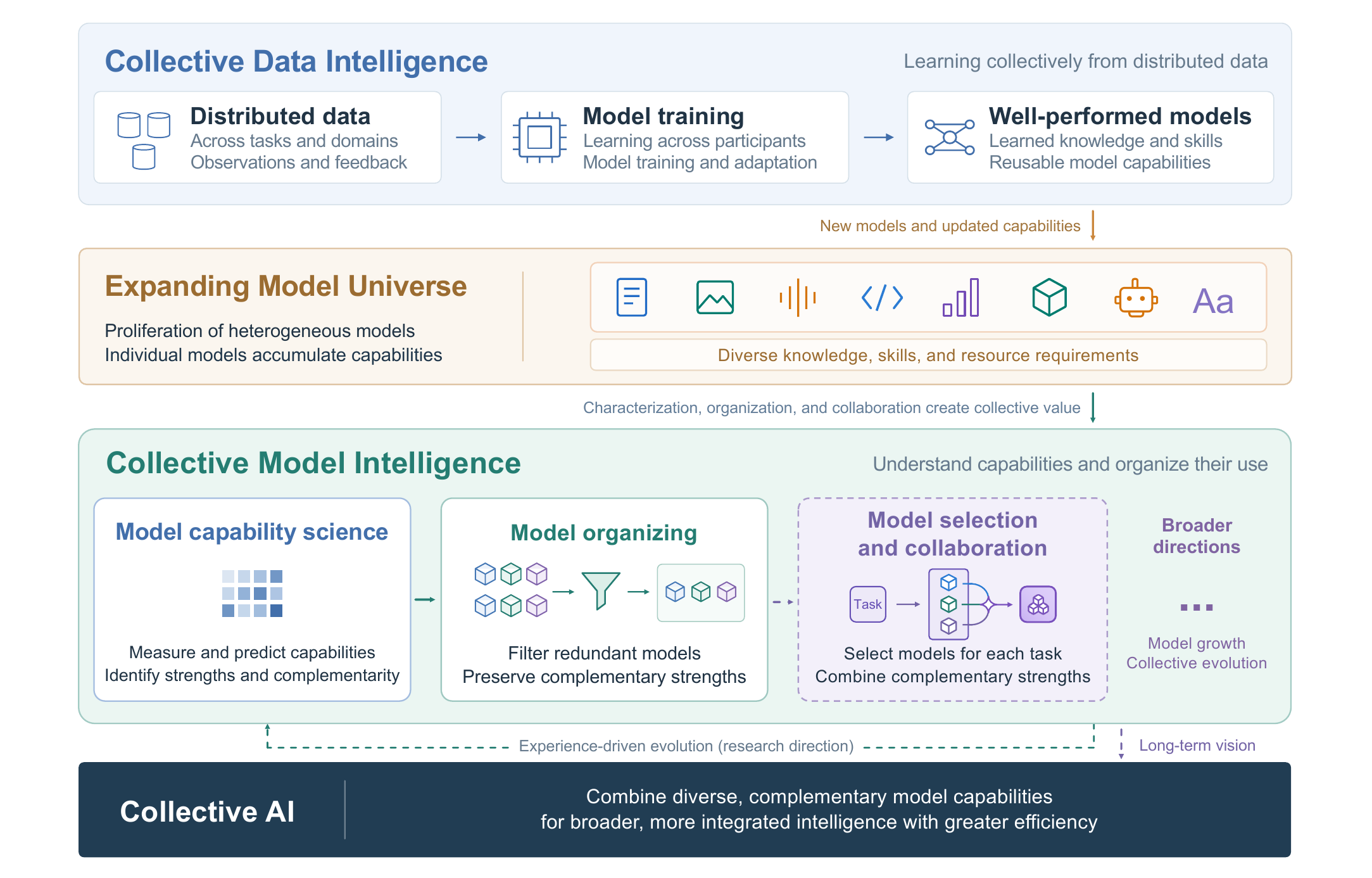}
\caption{Budgeted capability characterization by Model Eval. Mos query results
and a fixed reference snapshot guide profile updates and further queries.
Profiles retain evidence and uncertainty.}
\label{fig:model-evaluation}
\end{figure}

\paragraph{Evidence and the two prediction stages.}
At round $k$, $\mathcal H_k$ contains the accumulated query--score pairs
$(x,y_*(x))$ for $f_*$. It starts from any permitted, compatible scores in
$\mathcal H_0$ and grows as new queries are scored. Newly acquired scores
for $f_*$ do not enter the historical fitting set $\mathcal O_q$.
$\mathcal J_k$ indexes the selected anchor domains, and $B_k$ is the
remaining part of the total evaluation budget $B$. This budget uses the
study's declared unit, such as query count. At the framework level, let
$\mathcal Z_k=\{(j,\bar s_{*j},u_j):j\in\mathcal J_k\}$ denote the
anchor-domain identities, score estimates, and uncertainty.
For example, new coding-query results can refine a coding-domain
estimate, which then informs predictions for related domains.
Estimates below are
conditioned on the current round's evidence; round superscripts are shown
when comparing rounds.

The two stages have the following general form:
\begin{equation}
 \begin{aligned}
  \mathcal H_k
    &\xrightarrow{\text{within-domain estimation}} \mathcal Z_k,\\
  (S,\mathcal O_d,\mathcal Z_k)
    &\xrightarrow{\text{cross-domain reconstruction}}
      (\widehat{\mathbf s}_*,\boldsymbol\sigma_*).
 \end{aligned}
 \label{eq:evaluation-stage-interfaces}
\end{equation}
Under the fixed evaluation inventory and system snapshot, the within-domain
estimator maps accumulated query evidence to noisy scores for selected anchor
domains. The cross-domain reconstructor combines these scores with
historical model-domain evidence to estimate the full domain vector. Here
$u_j$ denotes the calibrated standard deviation of an anchor-score estimate,
while $\sigma_{*j}$ denotes calibrated prediction error for a completed score.
The estimators evaluated in this report produce point scores; calibrated
error estimates are an additional input to the risk-based allocation interface.
The graph-based and query-response methods are specified in
\cref{sec:sparse-anchor-reconstruction,sec:query-level}.

\paragraph{Budget allocation across the hierarchy.}
An action $a=(j,\mathcal U)$ evaluates a nonempty set of unscored queries
$\mathcal U\subseteq\mathcal Q_j$. Let $b_j(x)>0$ be the budget charged for obtaining
and scoring a response from $f_*$ to query $x$, measured in the same
declared unit as $B$. When query costs are additive, the action cost is
$b(a)=\sum_{x\in\mathcal U}b_j(x)$; the action is affordable when $b(a)\leq B_k$.
Under a query-count budget, each query costs
one unit, so $b(a)=|\mathcal U|$. Any batch-level charge is included in the total
action cost and must also be known at selection time. A new anchor domain starts
with a representative query set sufficient to estimate its score; an existing
anchor domain receives further query evaluations. Prescribed anchor domains and coverage
constrain the available actions. Without initial observations of $f_*$, the first
round prioritizes the declared domain coverage.

Let $w_j\geq0$, with $\sum_j w_j=1$, be domain weights fixed before
evaluation to reflect relative domain importance.
When applicable calibrated errors and action-conditioned future-risk
estimates are available, the whole-profile risk and expected reduction
following action $a$ are
\begin{equation}
 R_k=\sum\nolimits_{j=1}^{J}w_j\bigl(\sigma_{*j}^{(k)}\bigr)^2,\qquad
 \Delta_k(a)=R_k-\mathbb E[R_{k+1}\mid\mathcal H_k,a].
 \label{eq:evaluation-vector-risk}
\end{equation}
Here $\sigma_{*j}^{(k)}$ is the calibrated prediction error at round $k$.
The estimate of $\Delta_k(a)$ averages over
possible query outcomes predicted from current evidence, updating both
stages for each outcome. Among affordable, protocol-compatible actions
$\mathcal A_k$, choose
\begin{equation}
 a_k\in\operatorname*{arg\,max}\nolimits_{a\in\mathcal A_k}
               \Delta_k(a)/b(a).
 \label{eq:hierarchical-query-acquisition}
\end{equation}
Only actual scored responses are appended to $\mathcal H_k$; predicted
outcomes used to choose an action do not become observations. Deduct
$b(a_k)$ from $B_k$, rerun both stages on the accumulated evidence, and
continue with the updated profile. Expected outcomes may be approximated by
predictive sampling. The allocation uses calibrated error estimates, and
its computation counts toward total cost. With point-score estimates, use
a predeclared policy such as stratified random sampling.

\paragraph{Stopping, publication, and updates.}
Stop when no affordable action remains. When the calibrated risk rule is
available, also stop if a declared risk tolerance is met or the best expected
gain falls below a preset threshold. Model Eval
publishes a capability profile $\Omega_*$ that retains query observations,
anchor-score estimates, the completed domain-score vector, uncertainty, coverage,
selection trace, and
model, endpoint, task, scorer, and method versions.

Mio's organizing functions use domain scores for coverage checks and query
scores to distinguish
complementary successful cases in \cref{sec:capability-guided-organizing}.
These screening rules use complete measurements or calibrated capability
intervals. A profile refresh can change candidate
membership. Changing the model or
evaluation inventory starts a new snapshot.

Historical collection, fitting, and calibration are preparation costs.
Per-model evaluation costs cover query selection, model calls,
scoring, and both prediction stages.
\Cref{sec:efficient-model-evaluation} reports query-level
benchmark-score estimation from limited query observations and cross-domain
prediction from a small set of measured domain scores, using hidden outcomes
for verification. These studies assess within-domain estimation and
cross-domain reconstruction separately.

\subsection{Predicting Performance Across Domains}
\label{sec:sparse-anchor-reconstruction}
The cross-domain stage combines supplied anchor scores with the sparse
historical model-domain matrix to predict $f_*$'s full domain-score vector.
Observed entries form a bipartite graph $G$ of models and domains,
with no same-type edges. For each domain, compatible observed scores
determine a mean $\mu_j$ and stabilized standard deviation $\nu_j>0$.
Edge scores are standardized as $z_{ij}=(s_{ij}-\mu_j)/\nu_j$;
these statistics remain fixed throughout characterization.

\paragraph{Score-graph prediction.}
Each node starts with a trainable state $\mathbf g_v^{(0)}\in\mathbb R^{d_g}$.
With $\mathcal N(v)$ denoting its active neighbors, each layer updates
\begin{equation}
 \mathbf g_v^{(l+1)}=\operatorname{Update}_l\!\left(
 \mathbf g_v^{(l)},\;
 \operatorname*{mean}\nolimits_{u\in\mathcal N(v)}
 [\mathbf g_u^{(l)};\,z_{vu}\mathbf g_u^{(l)};\,z_{vu}]\right).
 \label{eq:domain-graph-update}
\end{equation}
Semicolons denote concatenation, $z_{vu}$ is the active edge score, and
empty-neighborhood means are zero. The update concatenates the node state
with the unweighted neighbor mean, score-weighted neighbor mean, and mean
edge score, then applies an affine map, GELU, dropout, a residual connection,
and layer normalization. Bidirectional propagation exchanges information
between models and domains. From the final states $\mathbf g_v$, a matrix
$W$ and scalar node biases $\beta_v$ decode unmeasured scores:
\begin{equation}
 \widehat z_{ij}
 =(W\mathbf g_{f_i})^\top\mathbf g_{d_j}/\sqrt{d_g}
   +\beta_{f_i}+\beta_{d_j},\qquad
 \widehat s_{ij}=\operatorname{clip}_{[0,1]}\!\left(\mu_j+\nu_j\widehat z_{ij}\right).
 \label{eq:domain-graph-decoder}
\end{equation}
Here $\operatorname{clip}_{[0,1]}(t)=\min\{1,\max\{0,t\}\}$ bounds
decoder predictions after inverse normalization; observed scores and supplied
anchor scores remain unchanged.
Historical fitting minimizes a Huber reconstruction loss on temporarily
masked observed edges, removed before message passing. Missing entries
are never training labels.

\paragraph{Conditioning on anchor scores.}
At round $k$, add temporary anchor edges $E_k$ between $f_*$ and
selected domains absent from the historical observation set:
\begin{equation}
 G_k=G\cup E_k,\qquad
 E_k=\{(f_*,d_j):j\in\mathcal J_k,\ (*,j)\notin\mathcal O_d\}.
 \label{eq:anchor-graph-conditioning}
\end{equation}
Each temporary edge has weight $\bar z_{*j}=(\bar s_{*j}-\mu_j)/\nu_j$.
With fitted parameters frozen, message passing on $G_k$ recomputes node
states and predictions. Later rounds replace these weights with updated
anchor-score estimates; they do not enter $\mathcal O_d$ or become training labels.

Compatible direct measurements take precedence in $\widehat{\mathbf s}_*$.
Other anchor-domain entries retain $\bar s_{*j}$, and the decoder fills
remaining entries with usable graph support. Query responses enter through
anchor scores, without duplicate query edges.

Absolute performance and relative specialization answer different questions.
Removing the decoder's model--domain interaction gives the reference score
$s_{ij}^0=\operatorname{clip}_{[0,1]}\!\left(\mu_j+\nu_j(\beta_{f_i}+\beta_{d_j})\right)$,
using the same clipping rule.
The specialization deviation is
\begin{equation}
 \delta_{ij}=\widehat s_{ij}-s_{ij}^0.
 \label{eq:capability-deviation}
\end{equation}
We identify relative strengths and weaknesses through practically meaningful
deviations supported by uncertainty in both the score and its reference.
Relative specialization and absolute competence use distinct references;
domain assessments rely on representative query coverage.

\paragraph{Extending the graph.}
A new model needs an initialized state, bias, and compatible anchor scores;
a new domain also needs a declared scoring protocol, normalization, and
historical support. Shared parameters may be reused, but new node parameters
require fitting or validated initialization. New nodes can alter existing
states, requiring new graph and calibration versions.

\subsection{Query-Level Performance Prediction}
\label{sec:query-level}
The within-domain estimation stage in
\cref{sec:hierarchical-anchor-estimation} uses historically fitted response
functions and observations of $f_*$ in $\mathcal H_k$. It predicts unscored
query outcomes and estimates full-domain scores $\bar s_{*j}$ for
$j\in\mathcal J_k$. The current cross-domain reconstructor in
\cref{sec:sparse-anchor-reconstruction} conditions on the anchor-score
estimates obtained in this stage.

\paragraph{Learning from historical responses.}
Let $\boldsymbol\theta_i\in\mathbb{R}^{d_q}$ represent model $f_i$'s latent ability,
$g_x$ a query-specific response function shared across models, and $p_x$ the likelihood
appropriate to that query's score type. Then
\begin{equation}
 y_i(x)\mid\boldsymbol\theta_i,g_x
   \sim p_x\!\left(\,\cdot\,\middle|g_x(\boldsymbol\theta_i)\right).
 \label{eq:query-response-model}
\end{equation}
Our evaluated estimator uses a conjunctive response function:
\begin{equation}
    g_x(\boldsymbol\theta)
    =
    \prod\nolimits_{h:Q_{xh}=1}
    \sigma(a_{xh}\theta_h-\beta_{xh}),
    \qquad a_{xh}>0,
    \label{eq:conjunctive-response}
\end{equation}
where $\sigma$ is the logistic function,
$Q_{xh}\in\{0,1\}$ indicates whether query $x$ involves
latent dimension $h$, and $a_{xh}$ and $\beta_{xh}$
are discrimination and bias parameters.
A model's ability vector links its query responses, allowing observations
to inform predictions for unscored queries.

\Cref{sec:efficient-model-evaluation} compares within-domain estimators
using limited-query benchmark-score estimation. Its separate cross-domain
experiment supplies measured domain scores as exact anchor scores and bypasses
query-response estimation. Learned ability coordinates summarize response patterns, not named semantic
capabilities. The domain graph has its own representation and loss.

\paragraph{Query prediction and domain aggregation.}
The evaluated estimator obtains $\boldsymbol\theta_*$
by applying the trained encoder to features constructed
from $\mathcal H_k$ and the permitted historical responses,
while keeping the query-specific response functions $g_x$ in
\cref{eq:conjunctive-response} fixed. The completed score
$\widehat y_*(x)$ preserves an observed score when available and otherwise
uses its predictive mean. The same rule applies to another model using
its own recorded evidence. Observed and predicted entries remain marked
in the profile.

For a domain with weighted-mean aggregation, the anchor-score estimate is
\begin{equation}
    \bar s_{*j}
    =
    \sum\nolimits_{x\in\mathcal Q_j}
    v_j(x)\widehat y_*(x),
    \qquad j\in\mathcal J_k.
    \label{eq:anchor-domain-estimate}
\end{equation}
The query weights satisfy $v_j(x)\geq0$ and
$\sum_{x\in\mathcal Q_j}v_j(x)=1$ within each domain.
They are distinct from the across-domain weights $w_j$ in
\cref{sec:hierarchical-anchor-estimation}.
The estimated score represents the full declared population, not the mean
of selected queries. For a non-mean rule, apply the declared aggregation
to joint predictive outcomes and take the conditional mean of the aggregate;
plugging means into a nonlinear metric is generally different.

Each anchor domain retains its query weights, coverage, selection design, and scoring
protocol. Observation coverage counts distinct scored queries in the
declared population. These anchor-score estimates feed cross-domain reconstruction in
\cref{sec:sparse-anchor-reconstruction}. As new query scores expand
$\mathcal H_k$, the within-domain estimator refreshes the anchor scores and
cross-domain reconstruction updates the profile used by the next round of
\cref{sec:hierarchical-anchor-estimation}.

\paragraph{Additional evidence and new populations.}
A fixed-observation study predicts from a supplied mask. An active study
can acquire more scores within the same matrices, changing the observation
set rather than the populations. Improvements due to extra calls must be
distinguished from improvements in the completion method. Uniform,
domain-concentrated, and outcome-dependent missingness are separate
conditions. The query-level method balances learned response-group
coverage and information about $\boldsymbol\theta_*$; those
groups need not coincide with semantic domains.

Within an existing query bank, a new model can fit its latent ability from
new scores while reusing response functions, subject to new-model
validation. A new query needs evidence to fit a response function or a
validated content-conditioned extension. A new domain also needs its
population, aggregation, and calibration before providing anchor scores.

\paragraph{Current implementation of the two stages.}
The within-domain stage uses a historically fitted query-response model
to infer $\boldsymbol\theta_*$ from $\mathcal H_k$ and aggregates predicted and
observed query outcomes under each anchor domain's declared scoring rule.
The cross-domain stage uses graph completion over the fixed historical
model-domain evidence, conditioned on the supplied anchor scores.
In the current implementations described in
\cref{sec:query-level,sec:sparse-anchor-reconstruction}, this framework-level
flow takes the following form:
\begin{equation}
 \begin{aligned}
  \mathcal Z_k
    &=\operatorname{Aggregate}_{\mathcal J_k}
      \!\left(\operatorname{Predict}_{\mathcal O_q}(\mathcal H_k)\right),\\
  \widehat{\mathbf s}_*
    &=\operatorname{GraphComplete}(S,\mathcal O_d,\mathcal Z_k).
 \end{aligned}
 \label{eq:implemented-evaluation-stages}
\end{equation}
Query prediction and domain aggregation produce anchor scores, and graph
completion returns the full domain-score vector.
\par

\section{Mio: Organizing Models for Selection and Collaboration}
\label{sec:mio}
Mio (Models in One) addresses model organization, selection and collaboration within
our system. It aims to turn complementary capabilities into better task
quality and more efficient resource use by maintaining a compact candidate
pool and coordinating its members under evolving task requirements and
budgets. Capability profiles from
\cref{sec:capability-characterization} guide which models to retain
and how to match them to the current task.

Capability evidence alone does not determine these decisions. Keeping only
the highest average performers can discard complementary specialists and
economical options, while comparing every available model increases selection
overhead. A useful candidate pool must preserve these contributions,
and request-level selection must match them to the current task state,
access requirements, and resource constraints. As a task progresses, both
its needs and its remaining budget can change, calling for a new choice.

Mio addresses these requirements in two stages, summarized in
\cref{fig:mio-method-overview}. \Cref{sec:capability-guided-organizing}
constructs and maintains a compact candidate pool
$\mathcal C\subseteq\mathcal F$ using domain coverage, query-level
complementarity, and service compatibility.
\Cref{sec:model-identification} develops model selection and collaboration
under the remaining cost and latency budgets, adapting model use and
handoff as task needs evolve. Model Space retains the full inventory, and Mos
executes the selected model. The reported experiments assess quality--cost
trade-offs and request-level robustness
(\cref{sec:empirical-evaluation}).

\begin{figure}[!htb]
\centering
\includegraphics[width=\linewidth]{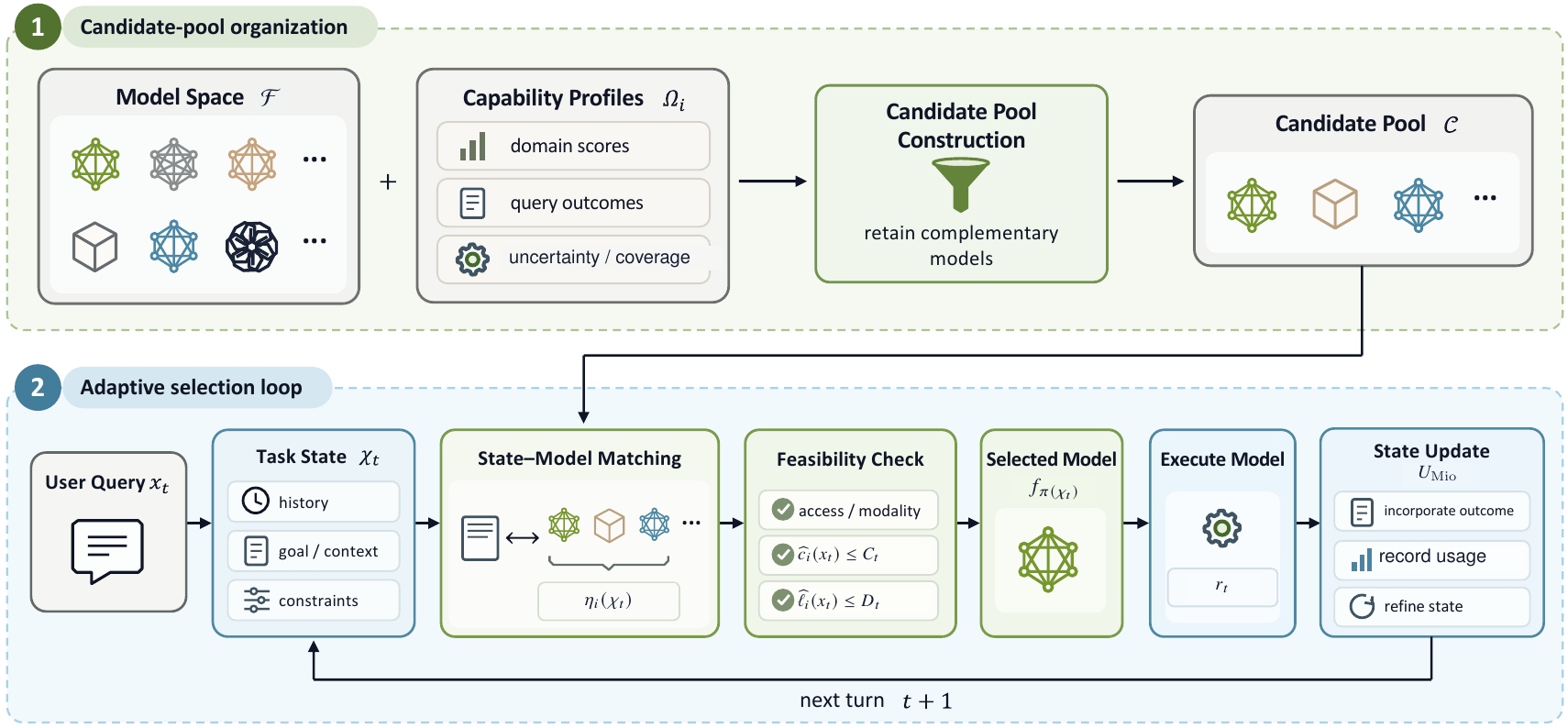}
\caption{Model organization, selection and collaboration in Mio. Model Eval profiles and
Model Space inform a complementary candidate pool; each decision selects a
feasible model for Mos execution and updates task context and resource budgets.
Capability-profile refresh follows separate scoring and evidence review by
Model Eval.}
\label{fig:mio-method-overview}
\end{figure}

\subsection{Capability-Guided Model Organizing}
\label{sec:capability-guided-organizing}
Organizing models aims to maintain a compact candidate pool while
preserving complementary strengths. Consider two models evaluated on the
same ten equally weighted queries in a domain: model A answers nine
correctly, while model B answers eight, including the one that A misses.
Keeping only A would discard a useful contribution from B, even though A
has the higher average score across the ten queries.

This example motivates a simple principle: remove a model only when
the available evidence indicates that a retained alternative can adequately
take over its useful role. These decisions are revisited as models,
capability evidence, or service conditions change.

Mio adapts the coverage-based learnware filtering idea
\citep{liu2025dynamicfiltering} to formalize this replacement principle.
Capability profiles and evaluation evidence from
\cref{sec:capability-characterization} support coverage and replacement-loss
checks, while current service information supports compatibility checks.

\paragraph{Supported and possible domains.}
Under a fixed domain scope $\mathcal D$ and evidence snapshot, we represent
each model--domain entry by a screening interval $[s_{ij}^{\mathrm{lo}},s_{ij}^{\mathrm{hi}}]$.
For profile predictions $\widehat s_{ij}\in[0,1]$ with applicable
calibration, define
\begin{equation}
 s_{ij}^{\mathrm{lo}}=\max\{0,\widehat s_{ij}-\kappa\sigma_{ij}\},\qquad
 s_{ij}^{\mathrm{hi}}=\min\{1,\widehat s_{ij}+\kappa\sigma_{ij}\}.
 \label{eq:admission-screening-bounds}
\end{equation}
The shared multiplier $\kappa\geq0$ controls the interval width. It can be
calibrated on held-out data, for example using a high quantile of
standardized absolute prediction errors. For complete, compatible
measurements, the interval reduces to $[s_{ij},s_{ij}]$, representing the
recorded score on a fixed query population. With neither complete
measurements nor applicable calibrated estimates, use $[0,1]$.
For competence thresholds
$\tau_j\in(0,1]$, define
\begin{equation}
 \mathcal S_i=\{d_j\in\mathcal D:s_{ij}^{\mathrm{lo}}\geq\tau_j\},\qquad
 \mathcal P_i=\{d_j\in\mathcal D:s_{ij}^{\mathrm{hi}}\geq\tau_j\}.
 \label{eq:admission-capability-sets}
\end{equation}
The set $\mathcal S_i$ contains domains supported by current evidence;
$\mathcal P_i$ also includes those where competence remains possible, so
$\mathcal S_i\subseteq\mathcal P_i$. Missing evidence keeps a domain in
$\mathcal P_i$ but not $\mathcal S_i$; an upper bound below $\tau_j$ excludes
it from both. Filtering operates on the qualified inventory
$\mathcal F_q=\{f_i\in\mathcal F:\mathcal S_i\neq\varnothing\}$,
whose models have at least one supported domain.

\paragraph{Directional replacement loss.}
To protect complementary successes, we assess the loss when $f_m$ replaces
$f_i$ using the completed query scores $\widehat y_i(x)$ from
\cref{sec:query-level}. For a domain with
weighted-mean scoring and applicable paired-loss calibration, define the
estimated loss $\widehat L_j(i,m)$ and the loss $L_j(i,m)$ used for filtering:
\begin{equation}
 \widehat L_j(i,m)=\sum\nolimits_{x\in\mathcal Q_j}v_j(x)
      [\widehat y_i(x)-\widehat y_m(x)]_+,\qquad
 L_j(i,m)=\widehat L_j(i,m)+\epsilon_{imj}.
 \label{eq:admission-estimated-query-loss}
\end{equation}
Here $[z]_+=\max\{z,0\}$. The query weights $v_j(x)$ follow the domain
aggregation in \cref{sec:query-level}. Gains on other queries do not cancel losses.
The margin $\epsilon_{imj}\geq0$ compensates for possible underestimation.
For complete, compatible paired measurements, use the observed scores
$y_i(x),y_m(x)$ and set $\epsilon_{imj}=0$. The screening value then reduces to
the recorded loss over the fixed query population:
\begin{equation}
 L_j(i,m)=\sum\nolimits_{x\in\mathcal Q_j}v_j(x)[y_i(x)-y_m(x)]_+.
 \label{eq:admission-query-loss}
\end{equation}
For binary scores, this is the weighted proportion of queries where $f_i$
succeeds and $f_m$ fails. With neither complete paired measurements nor
applicable calibrated estimates, use $L_j=+\infty$ to block replacement.
The margin $\epsilon_{imj}$ can be calibrated on held-out model pairs by comparing
estimated losses with losses from complete paired observations. A high
quantile of the underestimation errors is one possible choice.
Calibration data must match the evaluation conditions and remain separate
from predictor fitting and final testing.

Combining these checks with service compatibility, we define replacement
of $f_i$ by $f_m$ for distinct $f_i,f_m\in\mathcal F_q$ under the same
service snapshot:
\begin{equation}
 \operatorname{Replace}(i,m)=1
 \quad\Longleftrightarrow\quad
 \begin{cases}
  \mathcal P_i\subseteq\mathcal S_m,\quad
       \operatorname{Compat}(i,m)=1,\\
  L_j(i,m)\leq\epsilon_j,
       \quad\forall d_j\in\mathcal P_i.
 \end{cases}
 \label{eq:mio-replacement}
\end{equation}
Here $\epsilon_j\geq0$ is a finite, predeclared loss tolerance.
$\operatorname{Compat}(i,m)=1$ indicates that $f_m$ is verified to meet
$f_i$'s interface, modality, access, cost, and latency requirements under
the declared workload. Replacement is directional and need not be transitive.

\paragraph{Candidate retrieval and pool maintenance.}
To reduce comparison costs and keep the candidate pool up to date, Mio
combines coverage-based candidate retrieval with filtering triggered by
model, profile, or service changes.
We maintain a domain-to-model inverted index for
coverage-based candidate retrieval:
$\mathcal I_j(\mathcal C)=
\{f_m\in\mathcal C:d_j\in\mathcal S_m\}$. For a pool
$\mathcal C\subseteq\mathcal F_q$ and a newly qualified model
$f_{\mathrm{new}}\in\mathcal F_q\setminus\mathcal C$, index intersection
yields candidates to replace the newcomer:
\begin{equation}
 \bigcap\nolimits_{d_j\in\mathcal P_{\mathrm{new}}}\mathcal I_j(\mathcal C).
 \label{eq:mio-coverage-candidates}
\end{equation}
Conversely, candidates that the newcomer may replace are
$\{f_i\in\mathcal C:\mathcal P_i\subseteq\mathcal S_{\mathrm{new}}\}$.
Both sets are obtained using only the domain coverage condition.
Each candidate pair must also pass the query-level loss and service
compatibility checks in the corresponding direction:
$\operatorname{Replace}(\mathrm{new},m)=1$ for an existing model $f_m$ to
replace the newcomer, or $\operatorname{Replace}(i,\mathrm{new})=1$ for the
newcomer to replace an existing model $f_i$.

The pool-maintenance procedure uses this index to rerun filtering
whenever models arrive or profiles, domain scope, or service conditions
change. Initialize
$\mathcal C=\mathcal F_q$ from the updated qualified inventory, including
previously excluded models, reset replacement records, and build the index.
Process models in fixed identifier order. Before removing $f_i$, use index
intersection to find a direct replacement in $\mathcal C\setminus\{f_i\}$
for it and each excluded model assigned to it. Commit removal and update
records and the index together only if all affected models pass replacement
checks, preserving
\begin{equation}
 \forall f_i\in\mathcal F_q\setminus\mathcal C,\quad
 \exists f_m\in\mathcal C:\operatorname{Replace}(i,m)=1.
 \label{eq:mio-direct-replacement}
\end{equation}
Mio publishes the resulting pool and direct replacement records for model
selection, while Model Space retains all resources.

\subsection{Model Selection and Collaboration}
\label{sec:model-identification}
\label{sec:cost-aware-model-selection-collaboration}
Mio selects a suitable model from the organized pool $\mathcal C$ for the
current task state, subject to cost, latency, access, modality, and
execution-path constraints.

\paragraph{Budgeted routing objective.}
Consider a trajectory of $T$ sequential model calls. At step $t$, the policy observes the
bounded task state $\chi_t$, which includes the current request $x_t$, then
selects $f_{\pi(\chi_t)}$ from the candidate pool. Task-level utility may
depend on the complete execution trajectory, such as whether a coding task
ends with a correct patch. For budgeted routing, we use cumulative per-step
quality as a tractable surrogate:
\begin{equation}
 \begin{aligned}
 \pi^\star\in{}&\operatorname*{arg\,max}\nolimits_{\pi}\quad
    \mathbb E_{\pi}\!\Bigl[\sum\nolimits_{t=1}^{T}
      y_{\pi(\chi_t)}(x_t)\Bigr] \\
 \text{s.t.}\quad
   &f_{\pi(\chi_t)}\in\mathcal C\subseteq\mathcal F,\qquad
    \gamma_{\pi(\chi_t)}(\chi_t)=1,\quad t=1,\ldots,T,\\
   &\sum\nolimits_{t=1}^{T}c_{\pi(\chi_t)}(x_t)\leq C_0,\qquad
    \sum\nolimits_{t=1}^{T}\ell_{\pi(\chi_t)}(x_t)\leq D_0.
 \end{aligned}
 \label{eq:mio-routing-objective}
\end{equation}
Here $\gamma_i(\chi_t)\in\{0,1\}$ indicates whether $f_i$ passes the
current access, modality, interface, context-capacity, and execution-path
checks. The realized step-level quality score $y_i(x_t)$ is unavailable when
the decision is made; Mio must predict the corresponding ordering from
training evidence.
The resource terms $c_i(x_t)$ and $\ell_i(x_t)$ denote the total execution
charge and elapsed time for $f_i$ under the current service policy.
Here $C_0$ and $D_0$ are the initial trajectory budgets; $C_t$ and $D_t$
are the remaining budgets available before decision $t$, with
$C_1=C_0$ and $D_1=D_0$.
\Cref{eq:mio-routing-objective} specifies an idealized constrained objective
under realized resource consumption. The per-decision rule in
\cref{eq:model-selection} uses predicted consumption to approximate it.
The additive latency budget assumes sequential calls; concurrent workflows
would require a wall-clock accounting model.
Varying the initial budgets yields different quality--resource
operating points. A task-level extension can instead optimize an
application-defined trajectory utility under the same resource constraints.

This formulation exposes three coupled requirements. Performance is
request-dependent, so complementary specialists may outrank a stronger
average model. Cost, latency, and service compatibility determine which
actions are admissible. In a multi-step workflow, an executed result changes
both the information available to the next decision and the remaining
resources. For example, a coding agent may make successive model calls to
diagnose a bug, generate a patch, and revise it after running tests. Earlier
outputs and test feedback inform later requests, while incurred cost and
elapsed time reduce the budgets available for each new model choice.
A single-request task is the special case $T=1$. Rather than solve
the full trajectory online, Mio uses a receding decision loop: construct a
bounded task state, identify feasible candidates, score and select among them,
execute one model, and update the state.

\paragraph{Task-state construction.}
The latest request alone may omit information needed for routing, whereas an
unbounded transcript introduces irrelevant detail and growing decision cost.
Mio therefore constructs the bounded task state
\begin{equation}
 \chi_t=(o_t,x_t,\psi_t,\rho_t),
 \label{eq:mio-task-state}
\end{equation}
where $o_t$ is the persistent objective, $\psi_t$ is a compact execution
context, and $\rho_t$ contains current runtime requirements and observations.
The context retains application-visible conclusions, evidence, intermediate
artifacts, and unresolved requirements. The runtime component records modalities, context demand, access
constraints, prior usage, residual budgets $C_t,D_t$, and the service
information needed to interpret them. A task-state processor
initializes $\psi_1$ from the available context and updates it after each execution.
Throughout this section, $\pi(\chi_t)$ returns a model index and
$\mathcal C(\chi_t)$ denotes the feasible model set at the full task state.
Scores and resource terms also condition on the supplied context and service
snapshot; these fixed arguments are suppressed in $y_i(x_t)$, $c_i(x_t)$,
and $\ell_i(x_t)$.

\paragraph{Shared state--model selector.}
A shared state encoder and model-profile encoder, followed by projections
into a common space, represent the task and each candidate:
\begin{equation}
 \mathbf h_t=W_\chi E_\chi(\chi_t),\qquad
 \mathbf m_i=W_\Omega E_\Omega(\Omega_i,\mathbf e_i),
 \label{eq:mio-representations}
\end{equation}
where $W_\chi,W_\Omega$ are projection maps and $\mathbf e_i$ is an optional
learned embedding for residual properties
of a model in a fixed pool. The profile input uses scoped capability evidence
published by Model Eval; live endpoint state enters through $\rho_t$ and the
feasibility check.

The shared head scores each feasible candidate:
\begin{equation}
 \eta_i(\chi_t)=\operatorname{Head}\!\left[
  \mathbf h_t;\mathbf m_i;
  \mathbf h_t\odot\mathbf m_i;
  |\mathbf h_t-\mathbf m_i|\right].
 \label{eq:mio-interaction}
\end{equation}
Concatenation exposes the two representations; elementwise products and
absolute differences expose agreement and mismatch. The encoders and head
apply a shared scoring function across candidates. The pairwise objective
below trains $\eta_i(\chi_t)$ to reflect quality ordering for the current
state rather than global model strength.

\paragraph{Comparative selection learning.}
Informed by comparative model-conditioned ranking in EquiRouter
\citep{lai2026equirouter}, Mio learns state-conditioned quality orderings
from observed model outcomes.
Training trajectories supply states, and multiple candidates are executed
from the same state to obtain observed quality scores $y_i(x_t)$.
Sampled outcomes can be noisy \citep{lai2026dars}, so pairs are admitted only
when their observed quality difference exceeds a reliability margin
$\epsilon_y\geq0$. Define the quality-ordered pairs as
$\mathcal B_y(\chi_t)=\{(i,m):y_i(x_t)>y_m(x_t)+\epsilon_y\}$.
Mio minimizes
\begin{equation}
 \mathcal L_{\mathrm{rank}}(\chi_t)=
 \bigl(\max\{1,|\mathcal B_y(\chi_t)|\}\bigr)^{-1}
 \sum\nolimits_{(i,m)\in\mathcal B_y(\chi_t)}
 \log\!\left(1+\exp[-(\eta_i(\chi_t)-\eta_m(\chi_t))]\right),
 \label{eq:mio-ranking-loss}
\end{equation}
with an empty pair set contributing zero. The loss learns the ordering needed
at deployment without forcing heterogeneous task metrics onto one absolute
scale. For large pools, training pairs can be constructed from sampled
candidate subsets. The reliability margin is fixed in advance or selected
on development data.

\paragraph{Constraint-aware selection.}
For resource comparisons, define the development-normalized value of
$u\in\{c,\ell\}$ by
$\widetilde u_i(x_t)=(u_i(x_t)-u_{\min}^{\mathrm{dev}})/
(u_{\max}^{\mathrm{dev}}-u_{\min}^{\mathrm{dev}})$, using statistics fixed
on the development split (and zero when the range is degenerate). The
corresponding resource score is
\begin{equation}
 \zeta_i(x_t)=\lambda_c\widetilde c_i(x_t)
             +\lambda_\ell\widetilde\ell_i(x_t),
 \qquad \lambda_c,\lambda_\ell\geq0,\quad
        \lambda_c+\lambda_\ell=1.
 \label{eq:mio-resource-score}
\end{equation}
The weights are fixed in advance or selected on development data.

Mio screens candidates for the constraints in \cref{eq:mio-routing-objective}
using $\gamma_i(\chi_t)$, the residual budgets $C_t,D_t$, and the cost and
latency predicted under the
Mos scheduler's current endpoint policy. It first forms a per-decision feasible
set and, when that set is nonempty, applies the two-stage rule:
\begin{equation}
 \begin{aligned}
 \mathcal C(\chi_t)
   &=\{f_i\in\mathcal C:\gamma_i(\chi_t)=1,\
       \widehat c_i(x_t)\leq C_t,\
       \widehat\ell_i(x_t)\leq D_t\},\\
 \eta_t^{\max}
   &=\max\nolimits_{f_i\in\mathcal C(\chi_t)}\eta_i(\chi_t),\\
 \mathcal N_t
   &=\{f_i\in\mathcal C(\chi_t):
       \eta_i(\chi_t)\geq\eta_t^{\max}-\epsilon_\eta\},\\
 \pi(\chi_t)
   &\in\operatorname*{arg\,min}\nolimits_{i:\,f_i\in\mathcal N_t}
       \widehat\zeta_i(x_t).
 \end{aligned}
 \label{eq:model-selection}
\end{equation}
Here $\mathcal N_t$ is the feasible near-tie set, and $\epsilon_\eta$ is
calibrated on development data and fixed before test. The predicted resource
score $\widehat\zeta_i(x_t)$ applies \cref{eq:mio-resource-score} to
$\widehat c_i(x_t)$ and $\widehat\ell_i(x_t)$ using the same fixed
development normalization and weights.
Selection is therefore explicitly two-stage: predicted quality determines
the near-tie set, and resource use orders models only inside that set.
Predicted cost and latency also impose hard admissibility before either
stage. Changing $C_t$, $D_t$, access, or service conditions changes
$\mathcal C(\chi_t)$ without retraining the selector. This receding rule
provides a tractable per-decision approximation to
\cref{eq:mio-routing-objective}.

The estimates include routing overhead and the configured generation
allowance. The execution contract requires Mos
to recheck endpoint-level feasibility and enforce available usage caps. A model change triggers
a new Mio decision under the remaining budgets. If $\mathcal C(\chi_t)$ is
empty, Mio reports the unmet constraints.

\paragraph{Multi-step inference and model handoff.}
After selection, Mio passes the selected identity and residual budgets to
the Mos scheduler in \cref{sec:slo-aware-scheduling}. The chosen model produces
a result, and the feedback path in \cref{fig:mio-method-overview} advances
the state according to
\begin{equation}
 \begin{aligned}
 r_t&=f_{\pi(\chi_t)}(o_t,x_t,\psi_t),\\
 (\psi_{t+1},\rho_{t+1})
   &=U_{\mathrm{Mio}}\!\left(
       \psi_t,\rho_t,o_t,x_t,\pi(\chi_t),r_t,
       c_{\mathrm{exec}}(x_t),\ell(x_t)\right).
 \end{aligned}
 \label{eq:mio-state-transition}
\end{equation}
Here $c_{\mathrm{exec}}(x_t)$ and $\ell(x_t)$ are the realized charge and
elapsed time for the selected model in \cref{eq:mio-routing-objective}.
The update retains task-relevant information from $r_t$, records the selected
identity and observed usage, and deducts these amounts from the remaining
budgets. Model Eval incorporates scored results into a refreshed $\Omega_i$
under a compatible protocol with provenance and validity review, following
\cref{sec:model-eval}.
The next request or subproblem is then combined with this updated
context to form
$\chi_{t+1}$. Consequently, both task progress and resource consumption can
change the next feasible set and model ordering.

Continuing with the same model corresponds to
$\pi(\chi_t)=\pi(\chi_{t-1})$; a handoff occurs when the
identities differ. The receiving model is provided with the bounded context
$\psi_t$ and declared artifacts. Its access, interface, and context
requirements are checked again before invocation. Task decomposition and
termination remain application responsibilities, and each Mio decision
selects a single model.

\paragraph{Combining selection policies.}
Each participating router receives its declared request features, applicable
evidence, and the same feasible pool $\mathcal C(\chi_t)$. Routing overhead for
the combination is reserved before constructing that pool. Only routers
that can score every feasible candidate participate in the combination.

Let $\mathcal R$ be the active routers, each providing state-conditioned
quality scores. Convert each router's scores to descending ranks, using
average ranks for ties. The scores below express each router's ordering
on a shared $[0,1]$ scale. For a well-defined singleton case, we use
\begin{equation}
 \begin{aligned}
 \eta_{ri}(\chi_t)
   &=\begin{cases}
      1, & |\mathcal C(\chi_t)|=1,\\
      \bigl(|\mathcal C(\chi_t)|-\operatorname{rank}_r(f_i\mid\chi_t)\bigr)
            /\bigl(|\mathcal C(\chi_t)|-1\bigr), & |\mathcal C(\chi_t)|>1,
     \end{cases}\\
 \overline\eta_i(\chi_t)
   &=\sum\nolimits_{r\in\mathcal R}\alpha_r \eta_{ri}(\chi_t).
 \end{aligned}
 \label{eq:router-ensemble}
\end{equation}
Mio applies the two-stage rule in \cref{eq:model-selection},
using $\overline\eta_i(\chi_t)$ in place of $\eta_i(\chi_t)$.
The near-tie margin is calibrated on development data for the fused
rank scores, and resource use determines the final choice within
the resulting near-tie set.
Weights satisfy $\alpha_r\geq0$ and $\sum_r \alpha_r=1$. They are fixed in advance or fitted on
development data under a declared objective. Fixed weights can be
renormalized after omission; a learned combination needs a validated
fallback for the remaining routers. The combination requires
$\mathcal R\neq\varnothing$. Remaining ties are resolved by a fixed model
identifier.

\paragraph{Pool updates and computational cost.}
Mio records pool, profile, selector, and service versions. Candidate removals
can mask selector outputs. For newly admitted or updated models,
configurations using the identity embedding $\mathbf e_i$ need supporting
outcomes to learn or adapt that embedding. For configurations based solely
on capability profiles, pool updates include checking that the profile
encoder and selector apply to the new model's profile.

With fixed network widths, state encoding runs once and candidate scoring is
linear in $|\mathcal C(\chi_t)|$; exhaustive pair construction during training is
quadratic in pool size. Rank fusion additionally sorts each active router's
scores and combines them. These costs are separate from profile acquisition,
candidate-pool organization, and model invocation.

\paragraph{Relation to the RobustRouteBench evaluation.}
\Cref{sec:routestressbench} reports selector-level quality--cost performance
for a fixed pool and a single request ($T=1$) under monetary budgets.

\section{Mos: Model Serving OS for Scheduling and Operation}
\label{sec:model-serving}
Mos (Model Serving OS) provides shared execution for model evaluation and
application use. Model Eval, Mio, or the caller chooses the model; Mos
determines where and when its call executes. As shown in
\cref{fig:overview,fig:model-serving-workflow}, Mos invokes provider-managed
APIs or manages the Self-Managed Model Runtime Hub. Both paths share an
invocation contract, while model-file distribution and deployment apply
to the self-managed path. Endpoint scheduling preserves model identity;
model changes return control to Mio or the caller within remaining budgets.

\subsection{SLO-Aware Request Scheduling across Inference Endpoints}
\label{sec:slo-aware-scheduling}
An endpoint identifies a provider and serving configuration.
Endpoints exposing the same declared model can differ in prices, interfaces, rate
limits, latency, and availability, with these properties changing over time~\citep{li2026same}. Mos schedules calls among compatible endpoints
for the selected model.

The Mos scheduling implementation provides admission control, local capacity accounting, priority-based scheduling, pluggable window-based request reordering policies, SLO-aware assignment, Pareto-based optimal endpoint selection, and execution logging. Workload replay submits different types of requests through Mos according to their timestamps, allowing concurrent execution. Mos makes online scheduling decisions, while execution feedback updates capacity state and informs subsequent performance estimates. Sequential failover is supported over a bounded set of ranked endpoint candidates, subject to failure type and response progress. Optional preview simulations run in isolation from online capacity accounting.

\paragraph{Requests and constraints.}
A request \(x\) carries the selected model, invocation mode, input-token
count \(n_{\mathrm{in}}(x)\), estimated output length
\(\widehat n_{\mathrm{out}}(x)\), interface requirements, and priority.
Generation settings remain in the gateway request. Clients may specify SLO
targets; otherwise, the configured provider assigns default end-to-end (E2E) completion targets for
non-streaming requests and TTFT and TPOT targets for streaming requests.

The scheduler checks endpoint compatibility, health, cooldown, and remaining
local capacity. After further checking the SLO constraints, the scheduler proceeds to select a set of optimal endpoints with Pareto efficiency.

The scheduler uses separate timeouts for SLO computation, queue waiting,
and upstream calls. Cost estimates and E2E, TTFT, and TPOT targets guide
endpoint screening and ranking. The execution design in
\cref{sec:request-execution} additionally specifies upstream residual
allowances and shared budget and deadline enforcement across scheduling,
execution, and failover.

\paragraph{Endpoint state and availability.}
Endpoint configuration records supported models and interfaces, context
capacity, prices, requests per minute (RPM), tokens per minute (TPM),
and concurrency limits. Runtime state in Mos tracks the local reservations,
in-flight calls, health, and cooldown. In addition, our local endpoint runtime state tracker also keeps track of recorded and reported endpoint usage metrics. The bounded history retains outcomes, timestamps, usage, and available latency measurements.

Admission checks compatibility and health, then rechecks hard capacity before dispatch.
Capacity accounting applies the configured endpoint limits to requests
observed by this Mos instance.

\begin{figure}[htbp]
    \centering
    \includegraphics[
        width=\linewidth,
        trim={15bp 71bp 15bp 71bp},
        clip
    ]{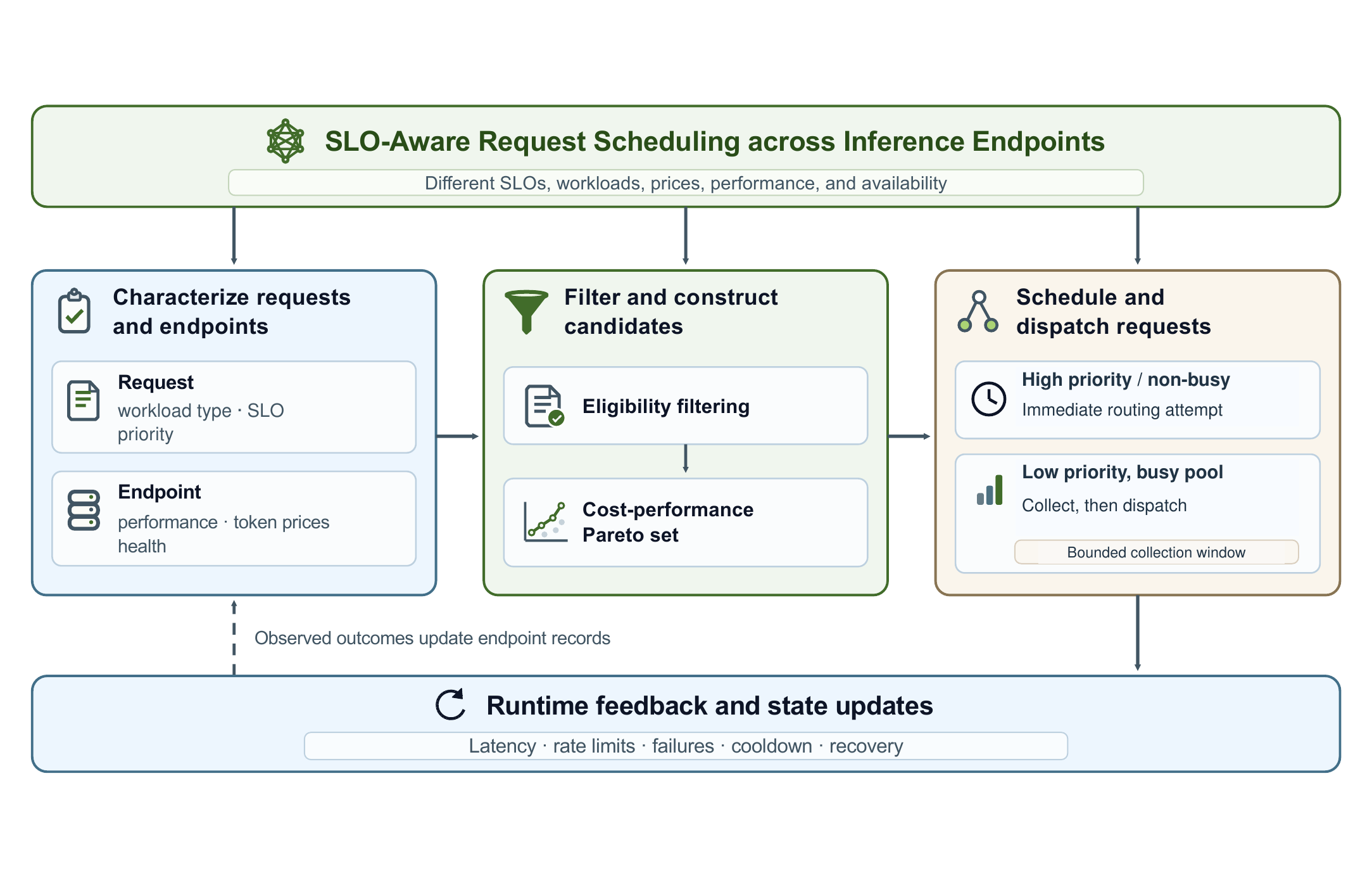}
    \caption{SLO-aware endpoint scheduling with model identity.
    High-priority requests bypass batch collection; low-priority requests enter
    a collection window when all eligible endpoints are busy. Endpoint selection
    combines capacity checks, SLO-aware screening, and Pareto-based ranking, with
    sequential failover over a bounded set of ranked candidates when the failure
    type and response progress permit.}
    \label{fig:slo-aware-endpoint-scheduling}
\end{figure}

\paragraph{Priority and collection windows.}
Load is assessed over compatible, healthy endpoints before applying hard
dispatch limits. An endpoint is busy when RPM or TPM utilization reaches
its configured threshold, or occupied concurrency reaches the maximum
minus reserved headroom. Reference defaults are 95\% for RPM and TPM and
three concurrency slots of headroom. Headroom must be configured below
the concurrency limit to avoid classifying an idle endpoint as busy.
The pool is busy when all eligible endpoints are busy; an empty pool has
no available service.

High-priority requests are scheduled over all hard-feasible endpoints immediately, while low-priority requests enter a short collection window when the endpoint pool is busy. The window is
released at its size or waiting-time limit. Released requests are ordered
by various policies such as weighted-length ordering, and dispatched one at a time, updating reservations
before deciding where to dispatch each subsequent request.

The default policy orders requests by weighted token length, using the
following score:
\begin{equation}
    s(x)=w_{\mathrm{in}}n_{\mathrm{in}}(x)
         +w_{\mathrm{out}}\widehat n_{\mathrm{out}}(x),
    \qquad w_{\mathrm{in}},w_{\mathrm{out}}\geq0,
    \label{eq:endpoint-window-order}
\end{equation}
with at least one positive weight. Token counts are used directly without
normalization. Ascending order favors shorter weighted lengths; ties
retain arrival order. The collection window bounds batch formation time.
The scheduling design applies residual deadlines to collection and subsequent
capacity waiting and rechecks feasibility at dispatch.

\paragraph{Endpoint selection.} For request
$x$ and endpoint $e$, estimated token charge is
\begin{equation}
    \widehat c_e(x)=
    \bigl(n_{\mathrm{in}}(x)c_e^{\mathrm{in}}
         +\widehat n_{\mathrm{out}}(x)c_e^{\mathrm{out}}\bigr)/10^6,
    \label{eq:endpoint-request-cost}
\end{equation}
where prices are per million tokens. For endpoint ranking, unavailable token prices are
estimated from the mean of known peer prices for the same model and endpoint
pool, with the estimation source recorded.

At decision time, endpoint performance is estimated from historical
observations matched by model, endpoint, invocation mode, and configured
token-length features. Recent and length-nearby observations receive greater
weight; the default statistic is a weighted mean. We also keep the state of whether the sample size is sufficient or not. An eligible candidate must have a valid estimated charge, sufficient observed
performance evidence, and satisfy the following constraints in our system settings.
\begin{equation}
    \hat L_{e,k}(x)\leq B_k(x),
    \qquad B_k(x)=m_{\mathrm{SLO}}T_k(x),
    \qquad 1\leq m_{\mathrm{SLO}}\leq2,
    \label{eq:endpoint-completion-feasibility}
\end{equation}
for each applicable timing metric \(k\), where \(\hat L_{e,k}(x)\) is the
predicted latency, \(T_k(x)\) is the assigned
SLO target and the default multiplier is \(1.2\) in our settings. Non-streaming requests
are screened on E2E; streaming requests are screened independently on TTFT
and TPOT. If the endpoint lacks any applicable targets, it is excluded.
The endpoints that pass this round of screening are considered regular candidates.

Among regular candidate endpoints, the selection policy forms a Pareto frontier
of estimated charge and latency loss. For non-streaming calls, latency loss
is the predicted E2E divided by the corresponding SLO target.
For streaming calls, it is a weighted sum of the TTFT and TPOT ratios to
their respective boundaries, with weights \(\eta\) and \(1-\eta\), where
the default \(\eta\) is \(0.5\). An endpoint is dominated when another has
no greater charge or loss and is strictly better in at least one dimension.
Non-dominated endpoints are ranked by a weighted sum of charge divided
by the median positive candidate charge and latency loss, using
nonnegative weights summing to one. A positive cost scale is declared
if all charges are zero. The lowest-scoring frontier endpoint is selected
as primary. By default, dominated candidates fill the remaining positions
in a ranked list of up to \(N\) endpoints, with \(N=3\). If no SLO-screened
endpoint qualifies, selection proceeds to a separately marked best-effort
tier, considering endpoints with limited performance history first, followed
by those without recorded performance measurements.
Ranked candidates support sequential failover when the failure type and response progress permit. If no candidate
can be selected, Mos returns a scheduling failure so that Mio can decide
how to proceed.

\paragraph{Execution feedback and recovery.}
The implemented feedback path updates the capacity, usage, history, failure,
and cooldown state of each endpoint maintained by Mos.
By default, Mos allows up to three attempts per request on each endpoint.
After these attempts fail, it sends the request to the next available
endpoint in the ranked list, if one exists. Mos reports complete failure
when all ranked endpoints have failed. Once a response or streaming
iterator is returned, Mos considers the handoff successful; failures
during subsequent streaming iteration are handled by the upstream
components that consume the stream.

\subsection{Efficient Model Distribution for Cloud AI Inference}
\label{sec:efficient-distribution}
Our system stores downloadable model files in cloud object storage for
self-managed inference. During system scale-out, each new inference instance must acquire
model files, load parameters, and initialize its backend. Concurrent
downloading can compete for network bandwidth while allocated GPUs wait for data.
For example, 20 instances reading a complete 40~GB model transfer about
800~GB of data in total. Instance-local caches support reuse on the same server but may not
help other instances placed elsewhere.

\paragraph{Shared cache architecture.}
\Cref{fig:cache-architecture} introduces a distributed cache orchestrated
by Fluid~\citep{gu2022fluid}. Cloud object storage provides reliable data persistence;
the cache holds recently used or prewarmed model files in memory or on
SSDs. Fluid manages cache runtimes and data preparation in Kubernetes;
a runtime such as Alluxio or Curvine serves the files to inference instances.
Shared reuse reduces repeated object-storage transfers; each instance
retrieves the model files it needs through the common cache service.

\begin{figure}[htbp]
\centering
\includegraphics[width=\linewidth,trim={13bp 48bp 9bp 53bp},clip]{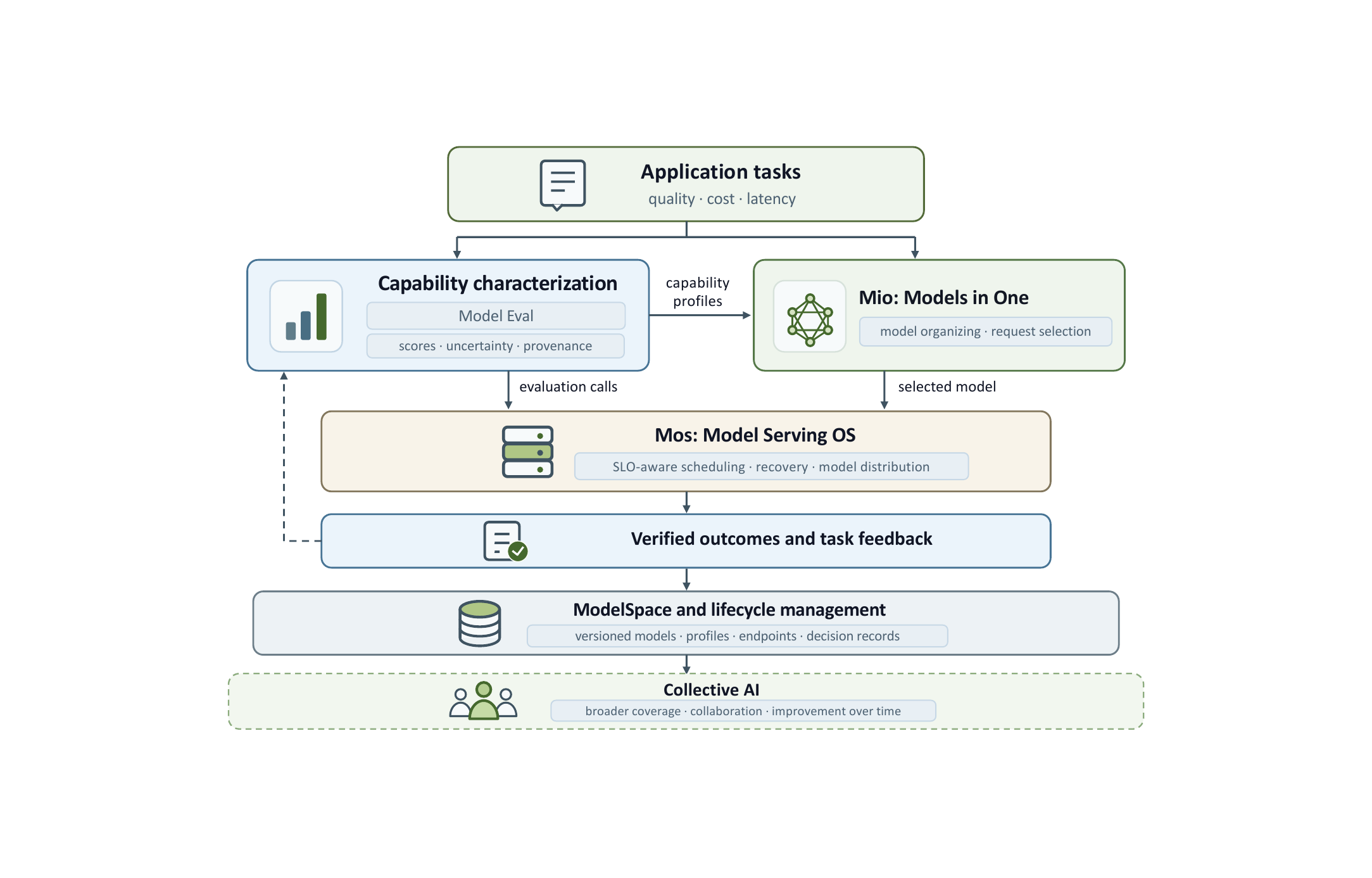}
\caption{Distributed elastic model cache orchestrated by Fluid.}
\label{fig:cache-architecture}
\end{figure}

The \emph{Smart Cache Managing and Scaling Service} provides three functions:
\begin{enumerate}[label=(\arabic*)]
  \item \textbf{Cache read/write.} Provides a common file-access path and
    publishes downloaded files only after integrity checks.
  \item \textbf{Traffic monitoring.} Records access frequency, cache hits,
    read bandwidth, and occupancy.
  \item \textbf{Policy adjustment.} Adjusts warmup, retention, eviction,
    and cache scaling policies using the access and occupancy records
    collected by traffic monitoring.
\end{enumerate}

\paragraph{Deployment and scaling.}
The platform checks access and the model revision and determines the required inference resources.
The cache service reuses verified cache hits or fetches missing files;
concurrent requests for the same data share acquisition.

Computation preparation and cache warmup can run in parallel. Once files are ready,
instances read them, load parameters onto GPUs, and pass readiness checks
before receiving traffic.

The cache-scaling design adds cache resources when capacity or read
bandwidth becomes limiting, then populates them with needed model files.
During sustained low load, runtime-supported scale-in preserves required
data and replicas. Cache resources and inference nodes scale separately,
with model preparation coordinated with instance startup.

\begingroup
\captionsetup{hypcap=false}
\section{Empirical Evaluation}
\label{sec:empirical-evaluation}
We evaluate Mio 1.0 Lite's quality--cost trade-offs, capability
characterization, request-level selection robustness on
RobustRouteBench, and model-file distribution.

\subsection{Mio 1.0 Lite Performance}
\label{sec:experimental-setup}
\label{sec:model-selection-collaboration-performance}
\label{sec:endpoint-scheduling-evaluation}
\begingroup
\captionsetup{hypcap=false}
\paragraph{Evaluation setup.}
We evaluate Mio 1.0 Lite on nine public benchmarks:
AutomationBench (public)~\citep{zapier2026automationbench},
PinchBench~\citep{pinchbench2026},
QwenClawBench~\citep{qwen2026qwenclawbench},
BFCL V3 Base Multi-Turn~\citep{mao2024bfclv3},
MT-Bench~\citep{zheng2023mtbench},
IFBench~\citep{pyatkin2025ifbench},
GPQA Diamond~\citep{rein2023gpqa},
Humanity's Last Exam (w/o tools)~\citep{phan2025hle}, and
AA-OmniScience (public)~\citep{artificialanalysis2025omniscience}.
The comparison includes eight fixed models and four routers: OpenRouter Auto
Router, its Beta variant, Jev Router, and vLLM Router.

Costs are mean fees per request (USD\,1 = CNY\,7); Mio sums fees of
models actually used in collaboration. Means equally weight nine 0--100 scores
(MT-Bench $\times 10$) before rounding.

\paragraph{Overall performance.}
Mio achieves the highest average score ($77.15/100$) and lowest average
cost (CNY\,0.0621 per request). It gains $6.35$ points over the
highest-scoring baseline, GPT 6.1 Sol, while reducing cost by $94.0\%$.
Against the cheapest baseline, Qwen 3.8 Flash, it gains $14.00$ points
and reduces cost by $16.4\%$.

\paragraph{Task-level results.}
Mio is non-dominated on all nine tasks and leads in score on
AutomationBench (public), PinchBench, BFCL V3 Base, IFBench, and GPQA
Diamond; BFCL and IFBench also have the lowest cost.
On QwenClawBench and AA-OmniScience (public), Mio stays within $0.43$
and $0.67$ percentage points of the leaders at $95.1\%$ and $85.3\%$
lower cost. On MT-Bench, Mio trades $0.223$ points (on 0--10) for
$89.7\%$ lower cost; on Humanity's Last Exam (w/o tools), it trades
$7.97$ percentage points for $97.5\%$ lower cost.

The \hyperlink{mio-frontpage-results}{opening overview} plots the
nine-benchmark means and three tasks; \cref{fig:mio-public-performance} shows
quality--cost scatter plots. \Cref{tab:mio-public-costs-tools,tab:mio-public-costs-knowledge,tab:mio-public-averages}
list all 117 score--cost pairs and the nine-benchmark averages.

\begin{figure}[!htbp]
\centering
\includegraphics[width=\linewidth]{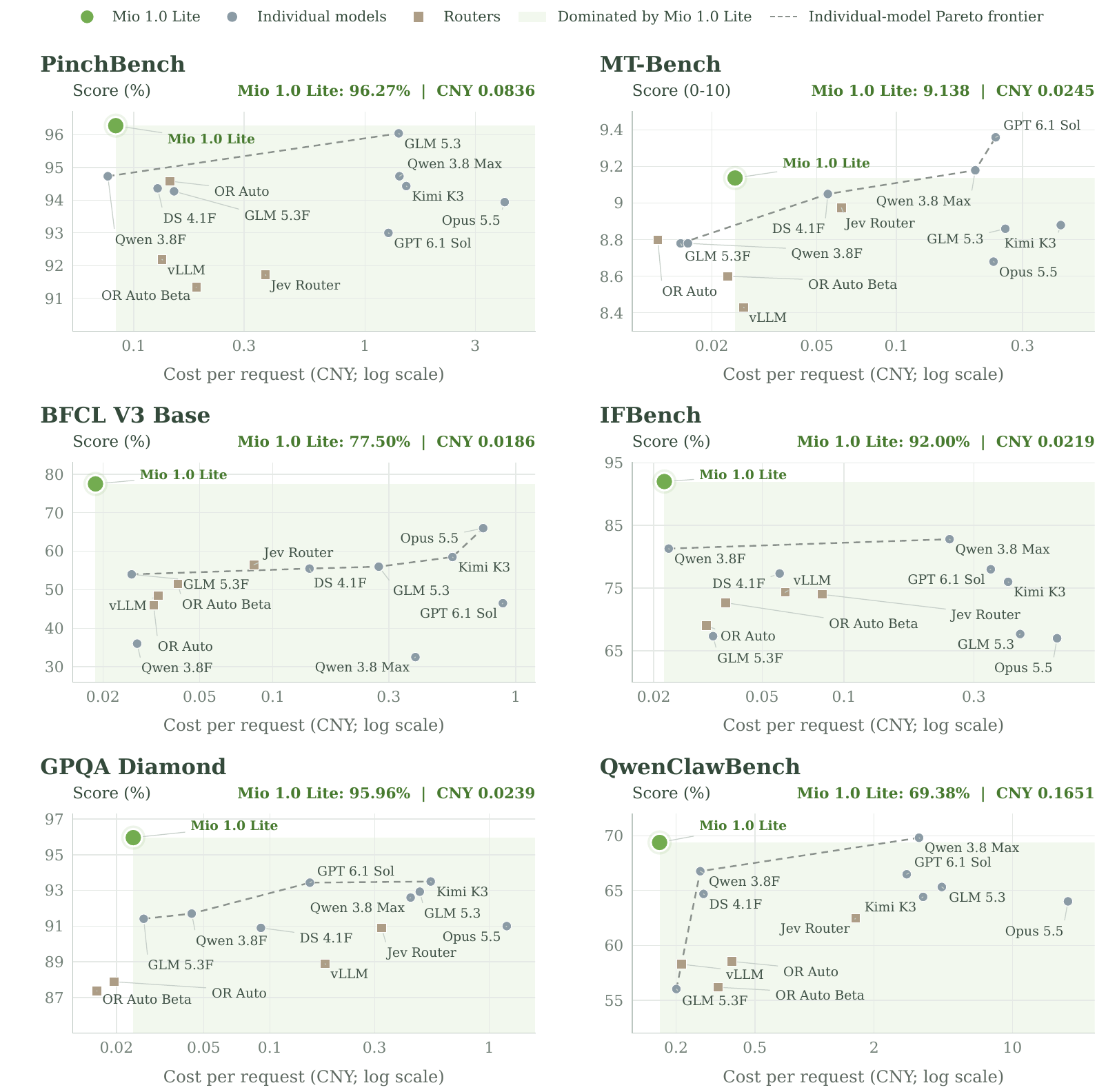}
\caption{Quality--cost trade-offs on six benchmarks.
Costs use logarithmic axes. Shading marks combinations dominated by Mio;
dashed lines connect observed individual-model Pareto points.}
\label{fig:mio-public-performance}
\label{fig:mio-public-tools}
\label{fig:mio-public-knowledge}
\end{figure}

\begin{table}[!p]
\begin{minipage}[t][\dimexpr\textheight-6pt\relax][s]{\linewidth}
\def\mioPublicTableGap{4pt plus 1fill}
\begingroup
\setlength{\parskip}{0pt}
\providecommand{\mioPublicTableGap}{4pt}
\captionsetup{skip=2pt,singlelinecheck=false}
\noindent\begin{minipage}{\linewidth}
\centering
\captionof{table}{Agentic execution and tool use. Scores are percentages; costs are CNY/request.}
\label{tab:mio-public-costs-tools}
\begingroup
\footnotesize
\setlength{\tabcolsep}{2pt}
\renewcommand{\arraystretch}{0.96}
\begin{tabularx}{\linewidth}{>{\raggedright\arraybackslash}p{0.32\linewidth}*{8}{>{\centering\arraybackslash}X}}
\toprule
\multirow{2}{*}[-6.35pt]{\textbf{Method}} & \multicolumn{2}{c}{{\bfseries\begin{tabular}[c]{@{}c@{}}Automation\end{tabular}}} & \multicolumn{2}{c}{{\bfseries\begin{tabular}[c]{@{}c@{}}PinchBench\end{tabular}}} & \multicolumn{2}{c}{{\bfseries\begin{tabular}[c]{@{}c@{}}QwenClaw\end{tabular}}} & \multicolumn{2}{c}{{\bfseries\begin{tabular}[c]{@{}c@{}}BFCL V3\end{tabular}}} \\
\cmidrule(lr){2-3} \cmidrule(lr){4-5} \cmidrule(lr){6-7} \cmidrule(lr){8-9}
 & \textcolor{ReportMuted}{\shortstack{Score\\(\%)}} & \textcolor{ReportMuted}{\shortstack{Cost\\(CNY)}} & \textcolor{ReportMuted}{\shortstack{Score\\(\%)}} & \textcolor{ReportMuted}{\shortstack{Cost\\(CNY)}} & \textcolor{ReportMuted}{\shortstack{Score\\(\%)}} & \textcolor{ReportMuted}{\shortstack{Cost\\(CNY)}} & \textcolor{ReportMuted}{\shortstack{Score\\(\%)}} & \textcolor{ReportMuted}{\shortstack{Cost\\(CNY)}} \\
\midrule
\rowcolor{BrandTint} \textbf{Mio 1.0 Lite} & \textbf{53.83} & \textbf{0.08} & \textbf{96.27} & \textbf{0.08} & 69.38 & \textbf{0.17} & \textbf{77.50} & \textbf{0.02} \\
DeepSeek V4.1 Flash & 49.50 & 0.58 & 94.36 & 0.13 & 64.69 & 0.28 & 55.50 & 0.14 \\
GLM 5.3 Flash & 48.80 & 0.13 & 94.27 & 0.15 & 56.02 & 0.20 & 54.00 & 0.03 \\
GLM 5.3 & 48.20 & 1.54 & 96.04 & 1.40 & 65.32 & 4.40 & 56.00 & 0.27 \\
Qwen 3.8 Flash & 49.33 & \textbf{0.08} & 94.73 & \textbf{0.08} & 66.76 & 0.26 & 36.00 & 0.03 \\
Qwen 3.8 Max & 27.30 & 0.91 & 94.73 & 1.41 & \textbf{69.81} & 3.37 & 32.50 & 0.39 \\
Kimi K3 & 30.80 & 1.89 & 94.43 & 1.51 & 64.42 & 3.54 & 58.50 & 0.55 \\
GPT 6.1 Sol & 48.00 & 2.89 & 93.00 & 1.26 & 66.48 & 2.92 & 46.50 & 0.88 \\
Opus 5.5 & 42.47 & 4.34 & 93.94 & 4.02 & 64.01 & 19.06 & 66.00 & 0.73 \\
\midrule
OpenRouter Auto Router & 45.50 & 0.10 & 94.58 & 0.14 & 58.54 & 0.38 & 46.00 & 0.03 \\
OpenRouter Auto Router (Beta) & 24.33 & 0.26 & 91.34 & 0.19 & 56.18 & 0.33 & 51.50 & 0.04 \\
vLLM Router & 44.83 & 0.24 & 92.19 & 0.13 & 58.28 & 0.21 & 48.50 & 0.03 \\
Jev Router & 49.83 & 0.30 & 91.72 & 0.37 & 62.46 & 1.61 & 56.50 & 0.08 \\
\bottomrule
\end{tabularx}
\endgroup
\end{minipage}
\par\vspace{\mioPublicTableGap}

\noindent\begin{minipage}{\linewidth}
\centering
\captionof{table}{Knowledge and reasoning. Scores are percentages; costs are CNY/request.}
\label{tab:mio-public-costs-knowledge}
\begingroup
\footnotesize
\setlength{\tabcolsep}{2pt}
\renewcommand{\arraystretch}{0.96}
\begin{tabularx}{\linewidth}{>{\raggedright\arraybackslash}p{0.32\linewidth}*{6}{>{\centering\arraybackslash}X}}
\toprule
\multirow{2}{*}[-6.35pt]{\textbf{Method}} & \multicolumn{2}{c}{{\bfseries\begin{tabular}[c]{@{}c@{}}GPQA\end{tabular}}} & \multicolumn{2}{c}{{\bfseries\begin{tabular}[c]{@{}c@{}}HLE\end{tabular}}} & \multicolumn{2}{c}{{\bfseries\begin{tabular}[c]{@{}c@{}}AA-OmniScience\end{tabular}}} \\
\cmidrule(lr){2-3} \cmidrule(lr){4-5} \cmidrule(lr){6-7}
 & \textcolor{ReportMuted}{\shortstack{Score\\(\%)}} & \textcolor{ReportMuted}{\shortstack{Cost\\(CNY)}} & \textcolor{ReportMuted}{\shortstack{Score\\(\%)}} & \textcolor{ReportMuted}{\shortstack{Cost\\(CNY)}} & \textcolor{ReportMuted}{\shortstack{Score\\(\%)}} & \textcolor{ReportMuted}{\shortstack{Cost\\(CNY)}} \\
\midrule
\rowcolor{BrandTint} \textbf{Mio 1.0 Lite} & \textbf{95.96} & \textbf{0.02} & 53.38 & 0.13 & 64.67 & \textbf{0.01} \\
DeepSeek V4.1 Flash & 90.90 & 0.09 & 39.10 & 0.25 & 54.50 & 0.03 \\
GLM 5.3 Flash & 91.41 & 0.03 & 36.75 & \textbf{0.09} & 40.67 & \textbf{0.01} \\
GLM 5.3 & 92.93 & 0.48 & 40.22 & 1.02 & 45.00 & 0.24 \\
Qwen 3.8 Flash & 91.70 & 0.04 & 35.90 & 0.13 & 24.83 & \textbf{0.01} \\
Qwen 3.8 Max & 92.60 & 0.44 & 43.60 & 1.31 & 39.33 & 0.23 \\
Kimi K3 & 93.50 & 0.54 & 43.50 & 1.93 & 50.17 & 0.46 \\
GPT 6.1 Sol & 93.43 & 0.15 & 52.92 & 0.54 & \textbf{65.33} & 0.10 \\
Opus 5.5 & 91.00 & 1.20 & \textbf{61.35} & 5.03 & 62.17 & 0.17 \\
\midrule
OpenRouter Auto Router & 87.88 & \textbf{0.02} & 38.51 & 0.24 & 45.83 & 0.03 \\
OpenRouter Auto Router (Beta) & 87.37 & \textbf{0.02} & 35.13 & 0.28 & 46.67 & 0.02 \\
vLLM Router & 88.89 & 0.18 & 30.44 & 0.15 & 29.17 & \textbf{0.01} \\
Jev Router & 90.91 & 0.32 & 47.91 & 1.07 & 45.33 & 0.08 \\
\bottomrule
\end{tabularx}
\endgroup
\end{minipage}
\par\vspace{\mioPublicTableGap}

\noindent\begin{minipage}{\linewidth}
\centering
\captionof{table}{Conversation and instruction following, with dataset-equal averages across all nine benchmarks. Scores use 0--10 for MT-Bench, percentages for IFBench, and 0--100 for the mean; costs are CNY/request.}
\label{tab:mio-public-averages}
\begingroup
\footnotesize
\setlength{\tabcolsep}{2pt}
\renewcommand{\arraystretch}{0.96}
\begin{tabularx}{\linewidth}{>{\raggedright\arraybackslash}p{0.32\linewidth}*{6}{>{\centering\arraybackslash}X}}
\toprule
\multirow{2}{*}[-6.35pt]{\textbf{Method}} & \multicolumn{2}{c}{{\bfseries\begin{tabular}[c]{@{}c@{}}MT-Bench\end{tabular}}} & \multicolumn{2}{c}{{\bfseries\begin{tabular}[c]{@{}c@{}}IFBench\end{tabular}}} & \multicolumn{2}{c}{{\bfseries\begin{tabular}[c]{@{}c@{}}Dataset-equal mean\end{tabular}}} \\
\cmidrule(lr){2-3} \cmidrule(lr){4-5} \cmidrule(lr){6-7}
 & \textcolor{ReportMuted}{\shortstack{Score\\(0--10)}} & \textcolor{ReportMuted}{\shortstack{Cost\\(CNY)}} & \textcolor{ReportMuted}{\shortstack{Score\\(\%)}} & \textcolor{ReportMuted}{\shortstack{Cost\\(CNY)}} & \textcolor{ReportMuted}{\shortstack{Score\\(0--100)}} & \textcolor{ReportMuted}{\shortstack{Cost\\(CNY)}} \\
\midrule
\rowcolor{BrandTint} \textbf{Mio 1.0 Lite} & 9.138 & 0.02 & \textbf{92.00} & \textbf{0.02} & \textbf{77.15} & \textbf{0.06} \\
DeepSeek V4.1 Flash & 9.050 & 0.06 & 77.33 & 0.06 & 68.49 & 0.18 \\
GLM 5.3 Flash & 8.780 & 0.02 & 67.33 & 0.03 & 64.12 & 0.07 \\
GLM 5.3 & 8.860 & 0.26 & 67.67 & 0.44 & 66.66 & 1.12 \\
Qwen 3.8 Flash & 8.780 & 0.02 & 81.30 & \textbf{0.02} & 63.15 & 0.07 \\
Qwen 3.8 Max & 9.180 & 0.20 & 82.80 & 0.24 & 63.83 & 0.94 \\
Kimi K3 & 8.880 & 0.42 & 76.00 & 0.40 & 66.68 & 1.25 \\
GPT 6.1 Sol & \textbf{9.360} & 0.24 & 78.00 & 0.35 & 70.81 & 1.04 \\
Opus 5.5 & 8.680 & 0.23 & 67.00 & 0.61 & 70.53 & 3.93 \\
\midrule
OpenRouter Auto Router & 8.800 & \textbf{0.01} & 69.00 & 0.03 & 63.76 & 0.11 \\
OpenRouter Auto Router (Beta) & 8.600 & 0.02 & 72.67 & 0.04 & 61.24 & 0.13 \\
vLLM Router & 8.430 & 0.03 & 74.33 & 0.06 & 61.21 & 0.12 \\
Jev Router & 8.975 & 0.06 & 74.00 & 0.08 & 67.60 & 0.44 \\
\bottomrule
\end{tabularx}
\endgroup
\end{minipage}
\par\vspace{4pt}

\noindent{\scriptsize\color{ReportMuted}Bold values mark the highest displayed score or lowest displayed cost in each group, including ties.}
\endgroup

\end{minipage}
\end{table}
\par
\endgroup

\subsection{Characterizing Models with Less Evaluation}
\label{sec:efficient-model-evaluation}
\label{sec:efficient-benchmark-evaluation}
\begingroup
\captionsetup{hypcap=false}
We evaluate the capability-characterization methods in \cref{sec:capability-characterization} through two studies: estimating benchmark scores from limited query observations, and predicting unmeasured domain scores from sparse historical evidence and a few measured anchor scores. Each study uses a fixed model inventory with separate historical models for predictor fitting and held-out models for score estimation.
The \href{https://github.com/SinapisAI/collective-model-intelligence/blob/main/docs/model-eval.md}{experimental documentation}
provides the detailed configurations, evaluation protocols, and implementations.
\paragraph{Query-level protocol.}
We compare our method, IRT \citep{polo2024tinybenchmarks},
and PSN-IRT \citep{zhou2026psnirt} using the query-response
interface in \cref{sec:query-level}, alongside Random and Historical mean. Our method combines the conjunctive response model with a learned ability encoder; one encoder pass estimates each evaluation model's ability vector. The three learned methods estimate each benchmark mean by uniformly
averaging observed query scores and predictions for unobserved queries.

Random samples uniformly from the full query pool at the same total budget and applies the sample mean to every benchmark. Historical mean uses each benchmark's visible historical responses. Within a run and visibility condition, Historical mean does not depend on the per-model query budget.

The data contain 25 models and 25,504 queries from 18 benchmarks. A fixed temporal split assigns 18 models to historical fitting and seven to evaluation. Five runs use nested masks with 1\% and 5\% historical visibility; hidden scores are excluded from fitting. Each evaluation model receives 5--40 scored queries across the pool. Under each condition, the learned methods receive identical observations from stratified sampling or a shared set of IRT-selected queries. Historical response functions remain frozen throughout evaluation.

MAE averages absolute errors against full-reference mean scores across the 18 benchmarks; reported values are averaged equally over five runs, seven models, and three query seeds.
In \cref{fig:query-anchors-comparison,fig:query-stratified-comparison},
our method achieves the lowest mean MAE at every displayed budget
under both historical visibility levels and selection settings.

\begin{figure}[!htbp]
\centering
\includegraphics[width=\linewidth]{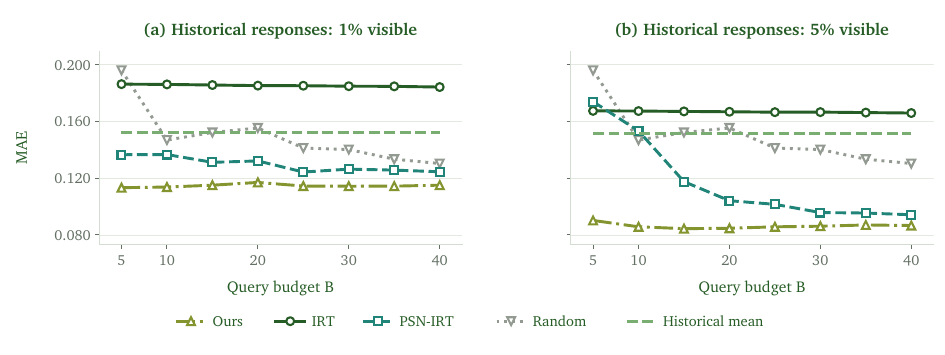}
\caption{Limited-query score estimation with shared IRT-selected
queries on held-out evaluation models. Lower MAE is better.}
\label{fig:query-anchors-comparison}
\end{figure}
\begin{figure}[!htbp]
\centering
\includegraphics[width=\linewidth]{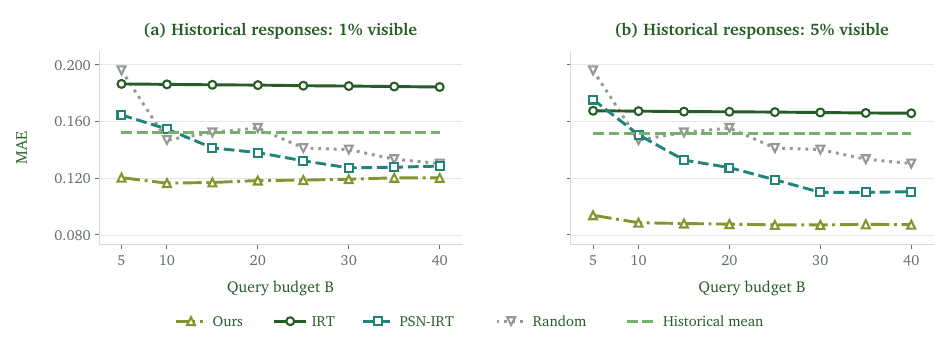}
\caption{Limited-query score estimation with stratified
sampling. Lower MAE is better.}
\label{fig:query-stratified-comparison}
\end{figure}

\label{page:query-evaluation-end}

\paragraph{Predicting unmeasured domain performance.}
This experiment compares the cross-domain predictor in
\cref{sec:sparse-anchor-reconstruction} with matrix-completion baselines. Each method receives the same sparse historical score matrix and measured anchor scores for each evaluation model, then predicts its remaining domain scores. Exact measured scores isolate cross-domain prediction from query-level estimation error.

\paragraph{Domain-level protocol.}
Behavioral clustering defines 32 evaluation domains, each with a fixed query population. One split assigns 30 models to historical fitting and five to evaluation. Domain clustering uses all 35 models, including the five evaluation models seen during development, defining transductive query populations. Four conditions hide 20\%, 40\%, 60\%, or 80\% of historical model--domain scores, using one fixed mask per condition. Each visible entry is the mean over its complete domain query population. Fitting uses the visible entries of the score matrix, and both observed scores and model predictions lie in $[0,1]$.

The budget $B\in\{0,1,2,4,8\}$ counts measured domains per evaluation model. Methods share nested, seeded anchor-domain sets, and measured scores are supplied as exact anchor scores using frozen historical standardization statistics. Historical parameters remain fixed during score prediction; unmeasured scores are used only for evaluation. Three fitting and three selection seeds yield 45 prediction vectors
per method, condition, and budget across the same five models.

\paragraph{Compared predictors.}
Graph is the two-layer score-graph implementation of
\cref{sec:sparse-anchor-reconstruction}. History mean uses observed historical domain means, and Mean + offset adjusts them using measured anchor scores. Raw Bias ALS combines additive biases with low-rank interactions in score space. The BenchPress core adaptation uses empirical biases and standardized logit-space ALS\citep{zeng2026benchpress}. PMF-MAP and
PMF-MCMC use probabilistic matrix factorization with fixed Gaussian
priors, motivated by performance prediction in
\citet{zhao2025crosspred}. MAP estimates factors by optimization; MCMC
conditions each evaluation model's factors on frozen historical posterior draws.
The BenchPress and PMF predictors independently adapt these core methods.

Graph uses two layers of width 16 and
100 training updates, with zero to four noisy anchor scores during training.
Evaluation on eight anchor domains extends beyond this training range.
Its residual updates, decoder, and masked-edge Huber objective follow
\cref{sec:sparse-anchor-reconstruction}. The factorization baselines use rank two; ALS
uses regularization 0.1 and a relative stopping tolerance of $10^{-6}$.
Each of PMF-MCMC's four chains uses 3,000 burn-in iterations and
2,000 retained draws.

\paragraph{MAE and budget dependence.}
MAE covers only unmeasured domains. Within each missingness
condition, it pools the five models and the fitting and selection
repetitions. The budget panel of \cref{fig:domain-performance-comparison}
averages the MAEs across the four missingness conditions with equal weight.
Displayed budgets are equally spaced, and both panels report mean MAE.
Graph has the lowest
average MAE at each nonzero displayed budget.
Its MAE decreases from 0.1679 at $B=1$ to 0.1474 at $B=8$.
At $B=8$, History mean gives 0.1597, Mean + offset 0.1590,
Raw Bias ALS 0.1670, the BenchPress adaptation 0.1615,
PMF-MAP 0.1628, and PMF-MCMC 0.1560. Graph's MAE is about
7.7\% lower than History mean in this setting.

\begin{figure}[!htbp]
\centering
\includegraphics[width=\linewidth]{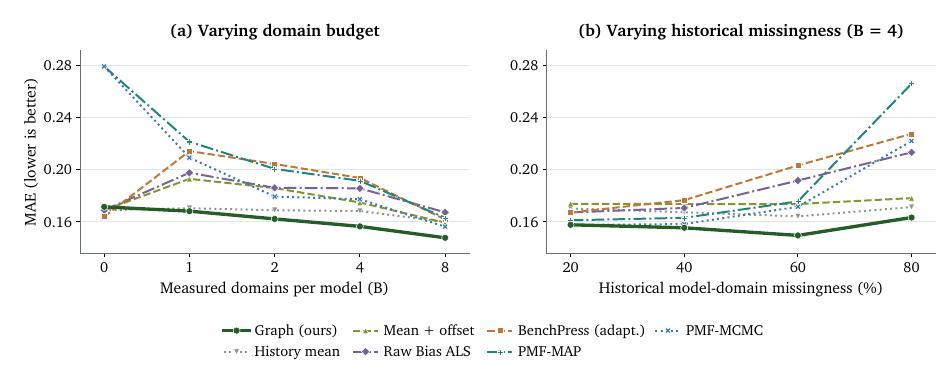}
\caption{Prediction MAE on unmeasured domains using shared
exact anchor scores. (a) Measured-domain budget, averaged equally over four
historical missingness levels. (b) Historical model--domain missingness at
$B=4$ measured domains per model. Lower MAE is better.}
\label{fig:domain-performance-comparison}
\end{figure}

\paragraph{Dependence on historical missingness.}
At $B=4$, within Graph's training range,
Graph has the lowest reported MAE at 40\%, 60\%, and 80\% missingness: 0.1551, 0.1493, and 0.1630.
At 20\%, Graph records 0.1574, close to PMF-MCMC's 0.1573. Each missingness rate uses one fixed mask.

\label{page:benchmark-evaluation-end}
\endgroup

\subsection{Model Selection Robustness on RobustRouteBench}
\label{sec:robustness-domains-modalities}
\label{sec:robust-model-selection}
\label{sec:routestressbench}
\begingroup
\captionsetup{hypcap=false}
We evaluate the fixed-pool, single-request setting ($T=1$) of the Mio
selector (\cref{sec:model-identification}) under monetary budgets
against 12 baselines on six Standard/Robust splits spanning General Text,
Multimodal, and Chinese (\cref{tab:routestressbench-main-composition}).

\paragraph{Evaluation setup.}
\label{sec:dynamic-ecosystems}
\label{sec:service-recovery-evaluation}
Methods use ORBIT's interfaces~\citep{lai2026orbit} with shared candidate
pools, training/development evidence, and frozen outcomes within each scenario.
Within each scenario, Mio and the baselines share candidate pools and
evaluation requests, with method-specific configurations. Multimodal training
and evaluation use independently constructed datasets.
The Multimodal training set is drawn from 18 MMR-Bench-V2
sub-benchmarks~\citep{ma2026mmrbench}.
Selectors cannot access held-out utilities, references, or realized costs.
Mio is trained separately per scenario and tuned on 500 development requests
(400 Standard, 100 Robust). Excluding all development source groups leaves
18,982 test requests. Training and final-test requests use disjoint
base-query/source-sample groups within the benchmark sources and task
domains. Model-side evidence is learned from benchmark
training observations. Utility follows source-task scorers,
with failures scored zero.
All Multimodal methods use late fusion of frozen CLIP ViT-B/32 features
\citep{radford2021clip}: full-text chunks and all request images are pooled
separately, normalized, and concatenated with equal modality weights.
Three fitted runs use
the same evaluation requests and candidate pool; General Text and Chinese
use one run. The
\href{https://huggingface.co/datasets/SinapisAI/RobustRouteBench}{dataset card}
provides full protocols and results.

\begin{table}[htbp]
\centering
\caption{RobustRouteBench candidate pools, training sizes, and final-test
populations after excluding all development source groups.}
\label{tab:routestressbench-main-composition}
\scriptsize
\setlength{\tabcolsep}{3.5pt}
\begin{tabularx}{\linewidth}{@{}>{\raggedright\arraybackslash}p{0.15\linewidth}rrrr>{\raggedright\arraybackslash}X@{}}
\toprule
Scenario & Models & Train & Standard test & Robust test & Robust changes \\
\midrule
General Text & 63 & 3,000 & 3,688 & 2,894 & Semantic rewrite, language variation, long-context noise \\
Multimodal & 45 & 4,000 & 4,600 & 1,900 & Text rewrite, context noise,\newline image variation, combined perturbation \\
Chinese & 10 & 7,590 & 3,900 & 2,000 & Semantic rewrite, English translation, prefix noise \\
\bottomrule
\end{tabularx}
\end{table}

\paragraph{Metrics.}
We evaluate 100 evenly spaced nominal budgets between the cheapest and most
expensive fixed-model mean costs on each reporting population. These
endpoints use held-out costs for evaluator-side budget construction and are
shared across methods; they are not training or development statistics.
Predicted costs determine feasibility. If no candidate meets the predicted
cost budget, the benchmark evaluator selects the model with the lowest
predicted cost and retains its measured outcome. Held-out outcomes
determine quality.

Normalized AUC (nAUC) is the trapezoidal
area under the cumulative-maximum quality curve, divided by the budget range;
macro nAUC averages the six split scores equally. Peak quality
($Q_{\mathrm{peak}}$) is the highest curve value. Curves contain only evaluated
selector points.

Multimodal runs share each reporting population's nominal
budget grid; utility and realized cost are averaged pointwise before taking
the cumulative maximum or computing any curve metric. Robust split metrics
use the pooled robust request population; each robustness type is evaluated
separately on its own budget range, so averaging type-level nAUC values need
not recover pooled Robust nAUC. Latency is auxiliary and does not enter the
selection objective or impose a hard constraint in this study. For both
quality metrics, higher values indicate better selector performance.

\paragraph{Results.}
Mio selector achieves the highest macro nAUC, 0.6360 versus MLP's 0.6183,
and leads General Text Robust and both Multimodal splits
(\cref{tab:routestressbench-overall-results}). Its largest gain over the best
baseline is on Multimodal Standard (0.7077 versus 0.6359). \Cref{tab:routestressbench-overall-results} reports all six split rankings.

\begin{table}[!htbp]
\centering
\caption{RobustRouteBench nAUC across all 13 implementations, ordered by macro nAUC.}
\label{tab:routestressbench-overall-results}
\fontsize{9}{11}\selectfont
\renewcommand{\arraystretch}{1.2}
\setlength{\tabcolsep}{3pt}
\begin{tabularx}{\linewidth}{>{\raggedright\arraybackslash}p{0.245\linewidth}*{6}{>{\centering\arraybackslash}X}>{\centering\arraybackslash}p{0.13\linewidth}}
\toprule
\multirow{2}{*}[-3.25pt]{\textbf{Method}} & \multicolumn{2}{c}{\textbf{General Text}} & \multicolumn{2}{c}{\textbf{Multimodal}} & \multicolumn{2}{c}{\textbf{Chinese}} & \multirow{2}{*}[-3.25pt]{\textbf{Macro nAUC}} \\
\cmidrule(lr){2-3}\cmidrule(lr){4-5}\cmidrule(lr){6-7}
 & \textcolor{ReportMuted}{Standard} & \textcolor{ReportMuted}{Robust} & \textcolor{ReportMuted}{Standard} & \textcolor{ReportMuted}{Robust} & \textcolor{ReportMuted}{Standard} & \textcolor{ReportMuted}{Robust} & \\
\midrule
\rowcolor{BrandTint} \textbf{Mio selector} & 0.5924 & \textbf{0.5328} & \textbf{0.7077} & \textbf{0.6128} & 0.7653 & 0.6051 & \textbf{0.6360} \\
MLP & 0.5884 & 0.5231 & 0.6299 & 0.5918 & 0.7656 & 0.6108 & 0.6183 \\
EmbedLLM-RouteJudge & 0.5863 & 0.5304 & 0.5510 & 0.5869 & 0.7629 & 0.6087 & 0.6044 \\
Avengers-Pro & 0.6022 & 0.2865 & 0.6239 & 0.5680 & 0.7660 & 0.6054 & 0.5753 \\
Eagle & \textbf{0.6045} & 0.3359 & 0.6095 & 0.5264 & 0.7642 & 0.5982 & 0.5731 \\
Avengers & 0.5770 & 0.2832 & 0.6064 & 0.5553 & 0.7597 & 0.5801 & 0.5603 \\
RouteLLM-MF & 0.5939 & 0.2891 & 0.5914 & 0.5439 & 0.7532 & 0.5724 & 0.5573 \\
kNN & 0.5853 & 0.2671 & 0.5846 & 0.4995 & 0.7598 & 0.5694 & 0.5443 \\
HybridLLM & 0.5684 & 0.2616 & 0.5385 & 0.5106 & 0.7555 & 0.5721 & 0.5344 \\
Causal RM-Softmax & 0.4969 & 0.1774 & 0.6217 & 0.5477 & 0.7404 & 0.5628 & 0.5245 \\
SVM Router & 0.5789 & 0.2611 & 0.4869 & 0.4714 & 0.7394 & 0.5499 & 0.5146 \\
MIRT & 0.2771 & 0.1133 & 0.6359 & 0.5638 & 0.7437 & 0.5060 & 0.4733 \\
GraphRouter & 0.2468 & 0.1110 & 0.5558 & 0.5038 & \textbf{0.7675} & \textbf{0.6163} & 0.4669 \\
\bottomrule
\end{tabularx}
\par\vspace{3pt}
\begin{minipage}{\linewidth}
\scriptsize\color{ReportMuted}
Macro nAUC is the unweighted mean of the six split-level nAUC values.
Bold scores mark the best value in each column. Multimodal entries are
computed from pointwise mean curves over three fitted runs.
\end{minipage}
\end{table}

Across robustness types, Mio ranks first in both peak quality and nAUC for
Text semantic rewrites and long-context noise, and for Multimodal text
rewrites, context noise, and image variation
(\cref{tab:routestressbench-variant-results}). Text language variation has
the highest peak but second-ranked nAUC; Multimodal combined perturbations
rank second in peak and third in nAUC.

\begin{table}[!htbp]
\centering
\caption{Mio selector scores and ranks among 13
implementations for each robustness type. Multimodal metrics are computed
after pointwise averaging of the three run curves.}
\label{tab:routestressbench-variant-results}
\fontsize{9}{11}\selectfont
\renewcommand{\arraystretch}{1.2}
\setlength{\tabcolsep}{5pt}
\begin{tabular}{@{}llrcrc@{}}
\toprule
Scenario & Robustness type & $Q_{\mathrm{peak}}$ & Rank & nAUC & Rank \\
\midrule
\multirow{3}{*}{General Text} & Semantic rewrite & 0.5124 & 1 & 0.4829 & 1 \\
 & Language variation & 0.4913 & 1 & 0.4487 & 2 \\
 & Long-context noise & 0.6747 & 1 & 0.6529 & 1 \\
\addlinespace[4pt]
\multirow{4}{*}{Multimodal} & Text rewrite & 0.6455 & 1 & 0.6438 & 1 \\
 & Context noise & 0.6219 & 1 & 0.6021 & 1 \\
 & Image variation & 0.6289 & 1 & 0.6247 & 1 \\
 & Combined perturbation & 0.6099 & 2 & 0.5827 & 3 \\
\addlinespace[4pt]
\multirow{3}{*}{Chinese} & Semantic rewrite & 0.6156 & 5 & 0.6086 & 4 \\
 & English translation & 0.5893 & 7 & 0.5802 & 5 \\
 & Prefix noise & 0.6375 & 7 & 0.6287 & 5 \\
\bottomrule
\end{tabular}
\end{table}

\endgroup

\subsection{Efficiency of Model Distribution}
\label{sec:deployment-efficiency}
This experiment evaluates the model-file acquisition stage introduced in
\cref{sec:efficient-distribution}. Durable model files remain in cloud object
storage, while a Fluid-orchestrated cache serves prewarmed copies to
inference instances. The comparison asks how this cache changes model-file
read performance when multiple instances load the same model.

\paragraph{Cluster setup and acquisition paths.}
Experiments were run on a local self-managed six-node Kubernetes cluster with 64~vCPUs and
256~GB of memory per node, connected by a 10~Gbps intranet. Every loading pod
had a 2~vCPU limit and read a complete model into its process. The two arms
used the same model files and resource allocation. In the OSS arm, each pod
read directly from Alibaba Cloud OSS through a shared public egress.
In our environment, the observed client download speed from OSS ranged from 33.1 to 53.5~MB/s.
In the cache arm, complete copies of the same files were served from
a prewarmed Fluid-orchestrated cache running Fluid 1.1.0-alpha with Curvine
0.5.2-alpha, deployed with one master and one worker using memory for caching.
The experiment covered
models of 0.62, 1.52, and 18.84~GB at requested concurrency levels
$N\in\{1,3,5,7,10\}$.

Each model--concurrency--source condition was run three times. We averaged the
recorded durations across the three runs before calculating speedup. This
averaging reduces the influence of run-to-run variation.

\paragraph{Model dataset.}
To represent common model-size tiers in the self-managed
model-distribution path while covering representative inference workloads, we
selected three models spanning OCR, speech-transcript normalization, and
multimodal reasoning. hayai-ocr-v2 (0.62~GB; 8 files)
combines a SigLIP2 vision encoder with a custom decoder to recognize Japanese, Chinese, Korean,
and English text in image crops. s1-mini (1.52~GB; 13 files) is a 0.6B-scale
text-normalization model for speech-to-text output, primarily in English. It
converts raw ASR transcripts into natural written text by handling fillers,
repetitions and corrections, punctuation and capitalization, and the
formatting of numbers, dates, times, monetary amounts, and email addresses.
Ornith-1.0-9B (18.84~GB; 18 files)
is a multimodal reasoning model for coding agents, designed for software engineering, codebase analysis, terminal
tasks, tool use, and visual content understanding; it accepts text, image, and video inputs.

\paragraph{Measured quantities.}
Each pod reports its read duration $T_{\mathrm{pod}}$. A batch is admitted only
when all $N$ requested pods produce validated records. Its completion time
$T_{\mathrm{batch}}$ is measured by the runner from the common release signal
until all validated pod results are collected. Mean loading time is the
arithmetic mean of $T_{\mathrm{pod}}$ within the batch. Batch speedup is
$T_{\mathrm{batch,OSS}}/T_{\mathrm{batch,Fluid}}$, and per-instance speedup is
$\overline{T}_{\mathrm{pod,OSS}}/\overline{T}_{\mathrm{pod,Fluid}}$.
The cache is prewarmed before timing starts. The measured stage ends with
file retrieval into host memory; GPU parameter loading and inference follow
this stage.

\paragraph{Model distribution time cost.}
\Cref{fig:fluid-batch-load-time} reports batch completion time for distribution
of the three models. Direct OSS access becomes progressively slower as concurrency rises:
from $N=1$ to $N=10$, batch completion time increases from 13.47 to 122.39~s
for hayai-ocr-v2, from 35.24 to 283.97~s for s1-mini, and from 430.89 to
4311.11~s for Ornith-1.0-9B. With Fluid, the corresponding ranges are
5.92--9.52~s, 13.55--19.12~s, and 157.48--193.37~s. The largest model
therefore remains slower in absolute terms, but its Fluid completion time
stays near three minutes across the tested concurrency levels instead of
rising above one hour under direct OSS access. For hayai-ocr-v2, batch
completion time under Fluid varies non-monotonically with concurrency.
This variation may reflect the greater relative influence of transient
bandwidth fluctuations and startup overhead on short reads, together with
the amortization of shared startup overhead as the number of concurrent
readers increases.

\begin{figure}[htbp]
\centering
\includegraphics[width=\linewidth,trim={0bp 22bp 0bp 0bp},clip]{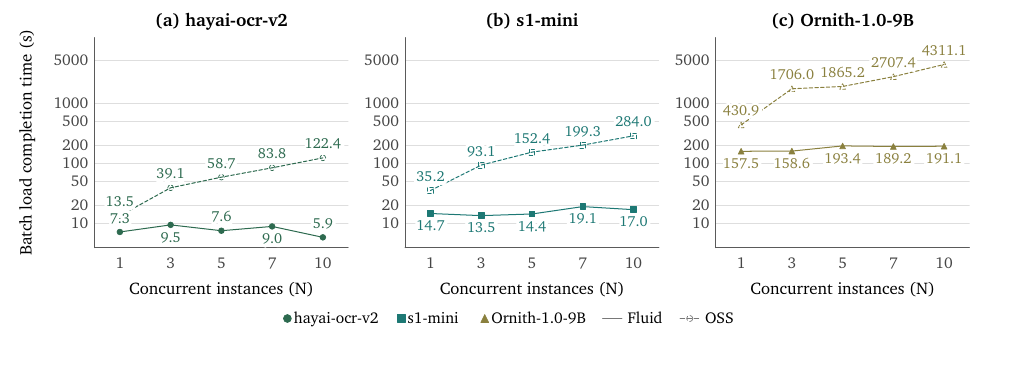}
\caption{Batch model-loading completion time: direct OSS versus a prewarmed
Fluid cache for three models. Solid: Fluid; dashed: OSS. Logarithmic vertical
axes; equally spaced concurrency levels.}
\label{fig:fluid-batch-load-time}
\end{figure}

\Cref{fig:fluid-per-instance-load-time} reports mean pod-read duration. The OSS means
increase from 13.47 to 110.96~s, 35.24 to 231.84~s, and 430.89 to
4242.59~s as concurrency rises from 1 to 10. The Fluid means remain within
5.58--7.47~s for hayai-ocr-v2, 12.97--14.69~s for s1-mini, and
157.00--185.17~s for Ornith-1.0-9B. Thus, the cache limits the growth in
model-loading time observed when more instances share direct OSS access.

\begin{figure}[htbp]
\centering
\includegraphics[width=\linewidth,trim={0bp 22bp 0bp 0bp},clip]{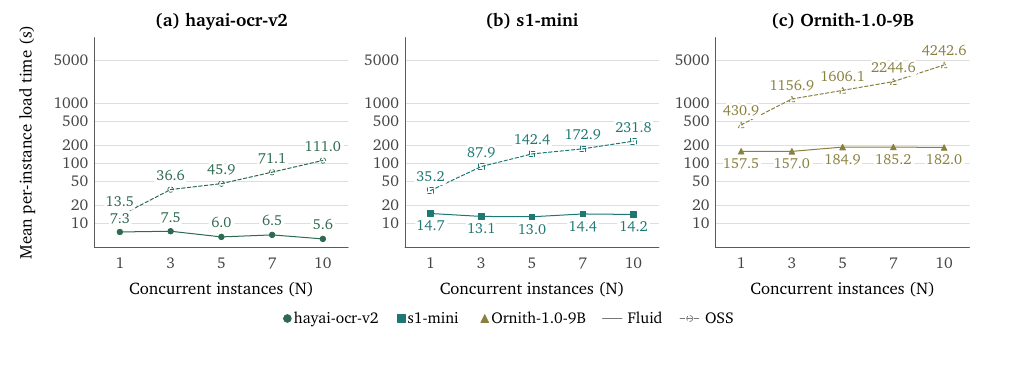}
\caption{Mean per-instance model-loading time for three models. Solid:
Fluid; dashed: OSS. Logarithmic vertical axes; equally spaced concurrency levels.}
\label{fig:fluid-per-instance-load-time}
\end{figure}

\begin{table}[!htbp]
\centering
\footnotesize
\captionsetup{hypcap=false}
\caption{Mean and peak Fluid speedup over direct OSS access.}
\label{tab:distribution-results}
\renewcommand{\arraystretch}{1.2}
\setlength{\tabcolsep}{5pt}
\begin{tabularx}{\linewidth}{@{}>{\raggedright\arraybackslash}X
                            >{\centering\arraybackslash}p{0.17\linewidth}
                            >{\centering\arraybackslash}p{0.17\linewidth}
                            >{\centering\arraybackslash}p{0.17\linewidth}
                            >{\centering\arraybackslash}p{0.17\linewidth}@{}}
\toprule
Model & Batch mean & Per-instance mean & Batch peak & Per-instance peak \\
\midrule
hayai-ocr-v2    & $8.73\times$  & $9.04\times$  & $20.67\times$ & $19.90\times$ \\
s1-mini         & $9.39\times$  & $9.68\times$  & $16.70\times$ & $16.36\times$ \\
Ornith-1.0-9B   & $12.00\times$ & $10.84\times$ & $22.56\times$ & $23.31\times$ \\
\bottomrule
\end{tabularx}
\par\smallskip
\raggedright\footnotesize\emph{Note.} Mean and peak summarize speedups
across the five concurrency levels. Ratios use durations averaged over three
runs per model, concurrency, and source configuration.
\end{table}

The summary in \cref{tab:distribution-results} condenses the detailed curves
into the magnitude of the observed time reduction. Mean batch speedup ranges
from $8.73\times$ to $12.00\times$ across the three models, and mean
per-instance speedup ranges from $9.04\times$ to $10.84\times$. Peak batch
and per-instance speedups reach $22.56\times$ and $23.31\times$,
respectively.

\par
\endgroup

\section{Conclusion and Future Vision}
\label{sec:conclusion-outlook}
A central implication of this work is that a model's value depends on its
role within a collective. Domain expertise, specialized skills, and
resource requirements shape where its strengths complement those of other
models. A system can draw on that complementarity to develop broader,
more integrated intelligence while using resources more efficiently. This report provides an
initial foundation for making such relationships measurable and usable,
and for advancing Collective AI through model capabilities accumulated across
developers and domains.

Our future work pursues two connected objectives. First, we aim to scale
the system to a much larger model population, including diverse vertical
models specialized for particular domains and tasks. Second, we aim to
enable experience-driven collective evolution through recursive
self-improvement of the system, using
its collective capabilities to diagnose weaknesses, revise its own
mechanisms, and validate the changes. Each cycle should strengthen both
task performance and the system's capacity for further improvement.

\par
\endgroup
\reportauthorsheading
\label{sec:contributions}
\begingroup\raggedright
\setlength{\parskip}{0pt} Within each group, names are listed in alphabetic order by last name.

\authorcontribution{Core Contributors}{\authorentry{Qun Dai}, \authorentry{Zhendong Du},
  \authorentry{Guannan Lai}, \authorentry{Hao-Yu Liu},
  \authorentry{Yi-Fan Ma}, \authorentry{Zi-Yu Mao},
  \authorentry{Tengzhong Sun}, \authorentry{Xin-Yi Zhang},
  \authorentry{Zi-Chen Zhao}}
\authorcontribution{Contributors}{\authorentry{Gelin Bian}, \authorentry{Zongyuan Cai},
  \authorentry{Ju Cang}, \authorentry{Zhangqin Chen},
  \authorentry{Shilei Gu}, \authorentry{Guoding Ji},
  \authorentry{Taojian Jiang}, \authorentry{Hao-Xuan Ma},
  \authorentry{Fan Wu}, \authorentry{Lin-Xing Wu},
  \authorentry{Tao Yu}, \authorentry{Xiao Zhang},
  \authorentry{Tianyi Zhou}}
\authorcontribution{Tech Lead}{\authorentry{Jian-Dong Liu}}
\authorcontribution{Academic Advisors}{\authorentry{Rong Gu}, \authorentry{Ming Li},
  \authorentry{Zhi-Hao Tan}, \authorentry{Han-Jia Ye},
  \authorentry{Yang Yu}, \authorentry{De-Chuan Zhan}}
\par\endgroup

\par
\phantomsection\addcontentsline{toc}{section}{References}
\bibliographystyle{learnware}
\AddToHook{cmd/bibsection/after}{\csdef{@currentlabel}{}\csdef{@currentlabelname}{References}\label{sec:references}}
\bibliography{references}
\end{document}